\documentclass[journal]{IEEEtran}
\ifCLASSINFOpdf
\else
\fi

\usepackage{stfloats}

\usepackage{url}
\usepackage{graphicx}

\usepackage{graphicx}
\usepackage{subcaption}
\usepackage{amsmath,amssymb} 
\usepackage{color}
\usepackage[hidelinks]{hyperref}
\usepackage{lipsum}
\usepackage{multirow}
\usepackage{array}
\usepackage{bm}

\usepackage{tabularx}
\newcolumntype{C}{>{\centering\arraybackslash}X} 
\usepackage{lipsum}

\newcommand{\ali}[1]{{\color{black}#1}}

\newcolumntype{U}{>
{\centering\let\newline\\\arraybackslash\hspace{0pt}}m{0.27\columnwidth}}

\begin{document}
\title{Deep Multi-Scale Feature Learning for Defocus Blur Estimation}
%
%
%

\author{Ali~Karaali,
~Naomi Harte,~and~Claudio~R.~Jung,~\IEEEmembership{Senior~Member,~IEEE}
\thanks{A. Karaali and N. Harte are with the School of Engineering, Trinity College Dublin. E-mail: (karaalia, nharte)@tcd.ie}
\thanks{C. Jung is with the Institute of Informatics, Federal University of Rio Grande do Sul. E-mail:crjung@inf.ufrgs.br}
}

\markboth{preprint for IEEE TRANSACTIONS ON IMAGE PROCESSING preprint}%
{Shell \MakeLowercase{\textit{et al.}}: Bare Demo of IEEEtran.cls for IEEE Journals}


\maketitle

\begin{abstract}
This paper presents an edge-based defocus blur estimation method from a single defocused image. We first distinguish edges that lie at depth discontinuities (called \textit{depth} edges, for which the blur estimate is ambiguous) from edges that lie at approximately constant depth regions (called \textit{pattern} edges, for which the blur estimate is well-defined). Then, we estimate the defocus blur amount at \textit{pattern} edges only, and explore an interpolation scheme based on guided filters that prevents data propagation across the detected \textit{depth} edges to obtain a dense blur map with well-defined object boundaries. Both tasks (edge classification and blur estimation) are performed by deep convolutional neural networks (CNNs) that share weights to learn meaningful local features from multi-scale patches centered at edge locations.
Experiments on naturally defocused images show that the proposed method presents qualitative and quantitative results that outperform state-of-the-art (SOTA) methods, with a good compromise between running time and accuracy.\end{abstract}


%
\IEEEpeerreviewmaketitle

\section{Introduction}
\label{sec:intro}

When a camera captures real 3D scenes, \ali{the 2D projection on the image plane tends to present defocused regions} due to the optical characteristics of lenses. Since the defocus blur amount depends on the distance of the captured objects to the focal plane, it generally varies from region to region in the image. This spatially varying blur is often represented by a defocus (blur) map that contains the size of the Circle of Confusion (CoC), which is typically characterized as a disc or Gaussian and described by a radius/scale parameter $C_{\sigma}$ or $\sigma$.

\begin{figure}[ht]
    \begin{center}
        \includegraphics[width=0.95\columnwidth]{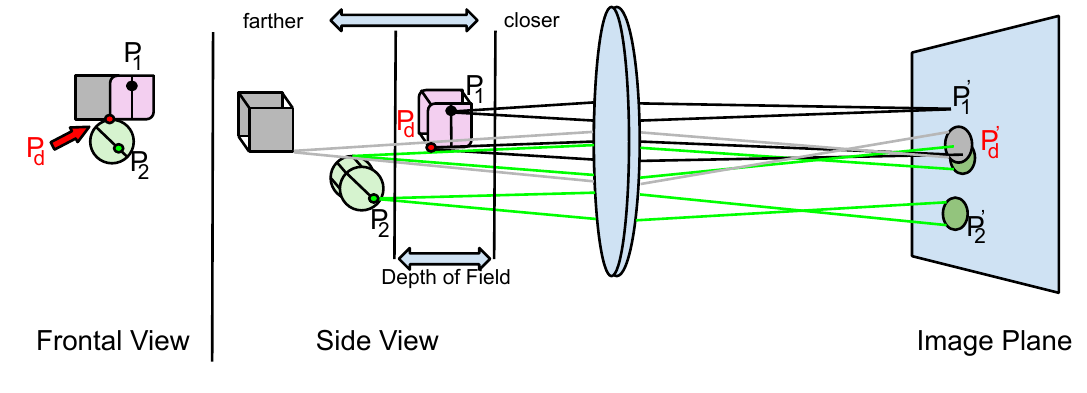}
         \vskip -0.250cm
        \caption{\label{fig:dofdesig} Pictorial representation of how Circle of Confusions (CoC) form from \textit{pattern} and \textit{depth} edges in a thin lens model.}
    \end{center}
     \vskip -0.50cm
\end{figure}

Estimating the local defocus blur amount in images has many potential applications, including depth estimation \cite{Gur_2019_CVPR}, in-focus object detection ~\cite{8537933,park2017unified}, salience region detection~\cite{UFO}, image retargeting~\cite{retargeting_our}, and non-blind deconvolution~\cite{Fortunato2014}. A typical categorization of existing defocus blur estimation methods is two-fold: edge-based and region-based methods. Edge-based methods use edge locations (which correspond to the high-frequency information and are more affected by blur) to estimate the local blur amount, then (optionally) propagate the defocus blur estimates from edge locations to the whole image. Region-based methods utilize local image patches and generally produce dense blur maps directly without any propagation scheme. In general, region-based methods are slower than edge-based methods since the defocus blur amount has to be estimated for each pixel location, while edge-based methods (initially) make estimates at only a fraction of the image pixels (i.e., the image edges)~\cite{OurTIP2018, karaaliPHD_dissertation}. 


The robustness of conventional defocus blur estimation methods~\cite{DefocusPaper,TIPpaper2012,jvcedgebased} is highly dependent on the strength and/or isolation level of edges, making these methods prone to error since natural images tend to present complex edges with multiple intensity and isolation levels. Also, the vast majority of existing edge-based defocus blur estimation methods follow a similar blur approximation model as presented in Pentland's work~\cite{Pentland87}, which relates the blur scale to the depth of an imaged 3D point based on a thin lens model.  
The main drawback of this formulation is that it is not valid for points that lie at depth discontinuities (e.g., boundaries between objects at different depth values), and the formulation should include a mixture of different blur parameters. Hence, assigning a single blur scale parameter to each edge point is ambiguous since the edge might be projected from a 3D point located at an object boundary with depth discontinuity. 
Fig.~\ref{fig:dofdesig} shows a pictorial representation of this behavior based on a thin lens model. Rays coming from an edge point formed by the object boundary ($P_d$) will be affected by other rays coming from other background (or foreground) objects, and the blur pattern will no longer be a simple circle. Instead, it will be formed by a combination of different circular patterns with different radii ($P_d'$ in Fig.~\ref{fig:dofdesig}). On the other hand, $P_1$ and $P_2$ are \textit{pattern} edges related to object texture, presenting a relatively constant depth neighborhood and generating a well-defined CoC on the sensor plane.

Our main contribution in this paper is the introduction of a Convolutional Neural Network (CNN) feature learning approach that jointly tackles: i) the
discrimination of \textit{depth edges} (i.e., edges that lie at depth discontinuities) from \textit{pattern edges} (i.e., edges that lie at relatively constant depth values); ii) multi-scale blur estimation for \textit{pattern edges} that uses input patches with varying sizes to account for different local edge patterns. As an additional contribution, we adapt a fast edge-aware guided filter to propagate blur information estimated at \textit{pattern edge} points to homogeneous regions, at the same time penalizing the propagation over \textit{depth edges}.  As will be discussed in the experimental results (Section~\ref{sec:experiments}), the final blur maps estimated by the proposed method for natural images yield competitive accuracy compared to all other SOTA methods in terms of the traditional error metric (MAE).

\section{Related Work}

Defocus blur estimation methods can be divided into two main groups: edge- and region-based methods. Also, deep learning strategies have been developed for this task in recent years.

\textbf{Edge-based methods} generally follow a common strategy: estimating the unknown defocus blur amount along image edges, obtaining a \textit{sparse blur map}, and then (optionally) propagating these blur estimates to the whole image via some interpolation/extrapolation methods to obtain a dense blur map (\textit{full blur map}). 
Pentland~\cite{Pentland87}, one of the pioneers of blur research, modeled a blurry edge by convolving a sharp edge with a Gaussian kernel. The standard deviation of the Gaussian kernel (i.e., the unknown blur scale) is then calculated using the intensity change rate along the edges. 
Elder and Zucker~\cite{elderandzucker98} proposed a simultaneous edge detection and blur estimation method. The method computes the blur scales measuring the zero-crossing of the third-order Gaussian derivatives along the gradient direction (using steerable filters). Both methods opted not to interpolate the blur estimates (e.g., estimated \textit{sparse blur maps} only). Later on, Zhuo and Sim~\cite{DefocusPaper} used the gradient magnitude ratio between the original and a re-blurred version (of the original image) to estimate the local blur amount at edge points. A bilateral filter was employed to smooth the sparse blur map, and alpha Laplacian Matting~\cite{LaplacianM} was chosen to propagate the estimated blur amounts to the rest of the image. The promising results of gradient magnitude usage (introduced in~\cite{DefocusPaper}) inspired many other researchers. For example,  the use of more than one re-blurring parameter was proposed in~\cite{fixedsigma1,karaalijungICIP2016} to deal with noise, while a multi-scale approach was explored by Karaali and Jung~\cite{OurTIP2018} to deal with edge-scale ambiguity. On the other hand, Chen et al.~\cite{icip2016fast} proposed a very fast \textit{full blur map} estimation method, using the same sparse estimation scheme as in~\cite{DefocusPaper} with a novel propagation scheme. They first over-segmented the image using superpixels~\cite{SLICpixel} and then computed an affinity matrix that measures the similarity between adjacent superpixels to assign a blur value to superpixels that do not overlap with image edges. Their method tends to produce piece-wise constant \textit{full blur maps}.

\textbf{Region-based methods} typically examine local image patches in order to estimate the unknown blur amount and they mostly employ a post-regularization scheme to produce visually coherent results. 
In the pioneering work of Chakrabarti et al.~\cite{CVPR2010FREQ}, the blur identification problem (both defocus and motion) was tackled in the frequency domain by exploring the \textit{convolution} theorem. They computed a likelihood function for a given candidate Point Spread Function (PSF), formulating a sub-band decomposition and Gaussian Scale Mixtures. This method later inspired Zhu et al.~\cite{ZhuFREQ}, who explored a continuous probability function to assess the unknown blur scale at each pixel location, analyzing the localized Fourier spectrum. Additionally, they incorporated color edge information and smoothness constraints to produce a coherent defocus blur map. Similar to~\cite{ZhuFREQ}, D’Andr\`{e}s et al.~\cite{TIPpaper20161} proposed the use of the localized Fourier spectrum, but they modeled the defocus blur estimation problem as image labeling. More specifically, they proposed labeling each image pixel with a discrete defocus blur scale using a machine learning method (regression tree fields -- RTFs), which provides a global consistency in the estimated defocus map.  Moreover, D’Andr\`{e}s et al. in~\cite{TIPpaper20161} created a naturally defocused image dataset with known (disk) defocus blur scales using a Lytro camera. Liu and colleagues~\cite{liu2020defocus} recently presented an extension of the RTF model by including edge information, which improved the results. 

\textbf{Deep learning} research has shown promising results in many areas ranging from image  classification~\cite{ILSVRC15}, 
to super-resolution~\cite{Li_2019_CVPR} and image deblurring~\cite{Gao_2019_CVPR}, to mention just a few
applications. In recent years, interesting deep strategies have also been developed for defocus blur estimation.
Zeng et al.~\cite{nnblur1} proposed a CNN architecture to learn meaningful local features in a superpixel level for blurry region estimation, and Zhang et al.~\cite{understandblur} explored the blur ``desirability'' in terms of image quality (at three levels  -- Good, OK and Bad) using a huge manually labeled dataset, which is not currently publicly available. On the other hand, Park et al.~\cite{park2017unified} proposed combining hand-crafted features with deep features to boost the performance following a similar strategy to edge-based methods (e.g., sparse blur map estimation along edges followed by an interpolation). Lee et al.~\cite{Lee_2019_CVPR} recently presented an end-to-end CNN architecture for the defocus blur estimation problem, which uses an additional \textit{Domain Adaptation}~\cite{domain} technique to transfer features from naturally defocused images to synthetically defocused images. Tang and colleagues~\cite{tang2020r2mrf} also presented an end-to-end network based on a series of residual refinements, but focusing on blur detection (i.e., separating in-focus from out-of-focus regions). 
In particular, end-to-end methods (such as~\cite{Lee_2019_CVPR,tang2020r2mrf}) can produce dense blur maps (as traditional region-based methods) while exploring tensor-based parallelism for fast execution, particularly when high-end GPUs are available. However, the performance tends to degrade as larger input images are used, as they might not fit entirely into the GPU memory.

In general, region-based approaches are costlier and more accurate than edge-based methods. In this paper, we present an edge-based method that outperforms existing edge- or region-based methods in terms of accuracy. The core idea of our method is to use only \textit{pattern} edge patches for defocus blur estimation to avoid CoC ambiguities at \textit{depth} edge points. 
For this purpose, our solution explores deep architectures that first distinguish \textit{pattern} from \textit{depth} edges using multi-scale image patches (edge classification task), then estimate the blur values along \textit{pattern} edges (blur estimation task), and finally  generates a full blur map by using a fast propagation scheme that respects  \textit{depth} edges. To the best of our knowledge, the issue of CoC ambiguity at depth edge points has only been dealt with by Liu et al.~\cite{TIP20162}, who model an edge point with two parameters (one parameter for each side of the edge). Although their method produces visually plausible results, using a two-sided blur model for depth edges is an oversimplification of the thin lens model since an edge point on a \textit{depth edge} might contain a mixture of different Circle of Confusion (CoC)~\cite{LensCOC,Lee_2019_CVPR}. Instead, we only use depth edges to prevent blur propagation across different objects.

\section{The Proposed Approach}
\label{sec:proposed}

\begin{figure*}[!t]
    \begin{center}   
        \includegraphics[width=\textwidth]{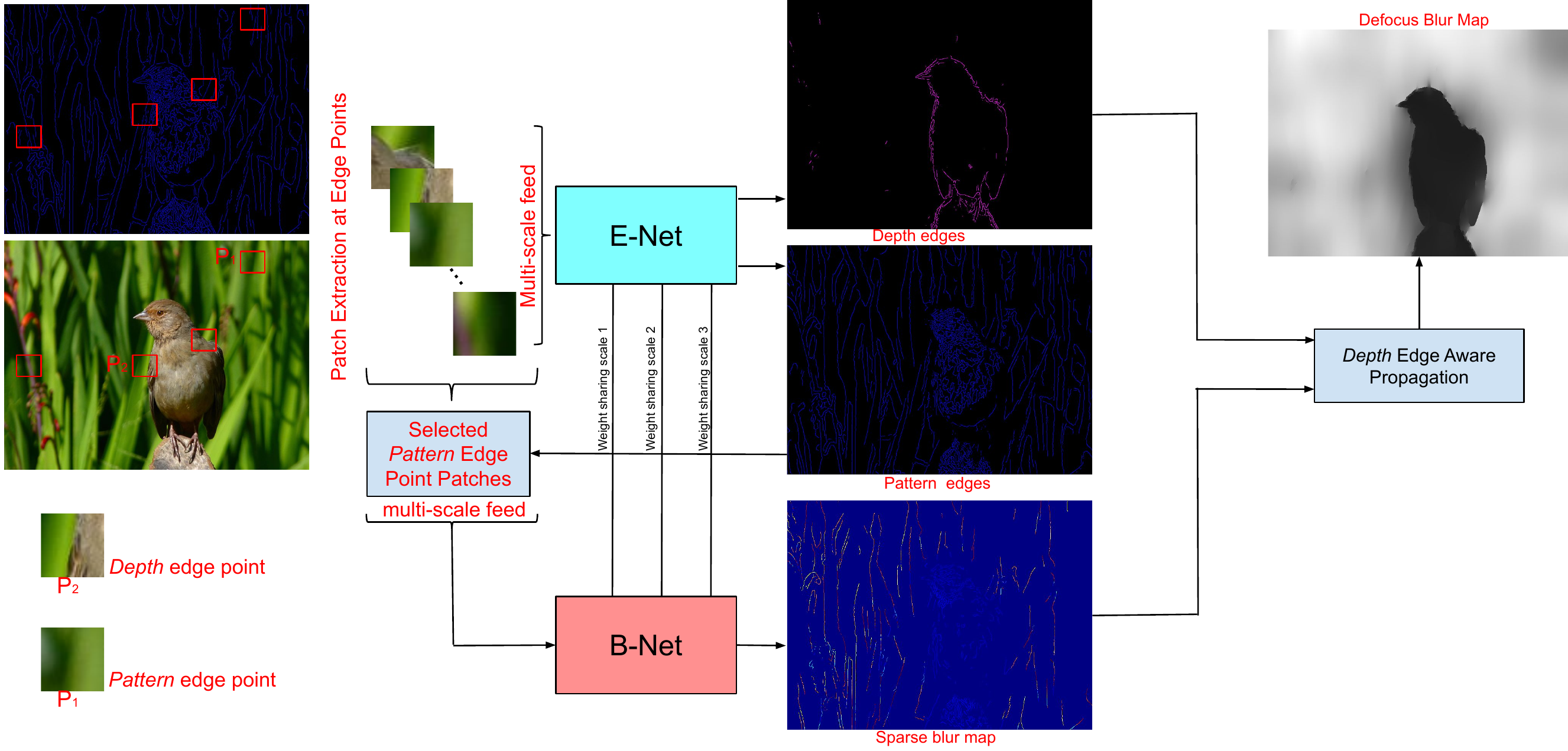}
         \vskip -0.250cm
        \caption{\label{fig:flowchart} Overview of our deep defocus blur estimation method.}
    \end{center}
 \vskip -0.750cm
\end{figure*}


Given a defocused input image $I^B$, our algorithm starts by computing an edge map of the input image. In this work, we used the well-known Canny detector~\cite{canny_ref}, but any other edge detector can be used.  Multi-scale image patches centered at the estimated edge locations are then fed to the first Convolutional Neural Network, called \textit{\textbf{E-NET}}, which classifies an edge as \textit{depth}- or \textit{pattern}-related. 
 The next step is to feed only patches related to \textit{pattern} edges to another Convolutional Neural Network, called \textit{\textbf{B-NET}}, which estimates the unknown defocus blur amount for a given edge point.  Finally, a fast image-guided filter that propagates the sparse blur estimates to the whole image while penalizing propagation over \textit{depth} edge points (which are related to object boundaries) is applied to obtain the final dense map. Fig.~ \ref{fig:flowchart} presents an overview of our method.

Although the \textit{pattern} vs. \textit{depth} edge classification problem is related to the 3D geometry of the scene, our method does not explore structural information as in single-image stereo approaches, such as~\cite{eigen2014depth}. Instead, we explore local blur information caused by defocus, which typically occurs when shallow Depth-of-Field (DoF) cameras are used. As such, both tasks (edge classification and blur estimation) are expected to share low-level features, which suggests some kind of communication between the two networks. Since blur estimation is a consolidated problem with existing annotated datasets (unlike the proposed edge classification task), we hypothesize that the low-level features learned in \textit{\textbf{B-NET}} can leverage the results of \textit{\textbf{E-NET}}. \ali{In fact, our initial tests explored one distinct network for each task (no weight sharing) and also a single network that branches off into the two tasks (full sharing of initial layers), but our partial weight sharing, as described next, presented the best results.}

\begin{figure*}[!t]
    \begin{center}
        \includegraphics[width=\textwidth]{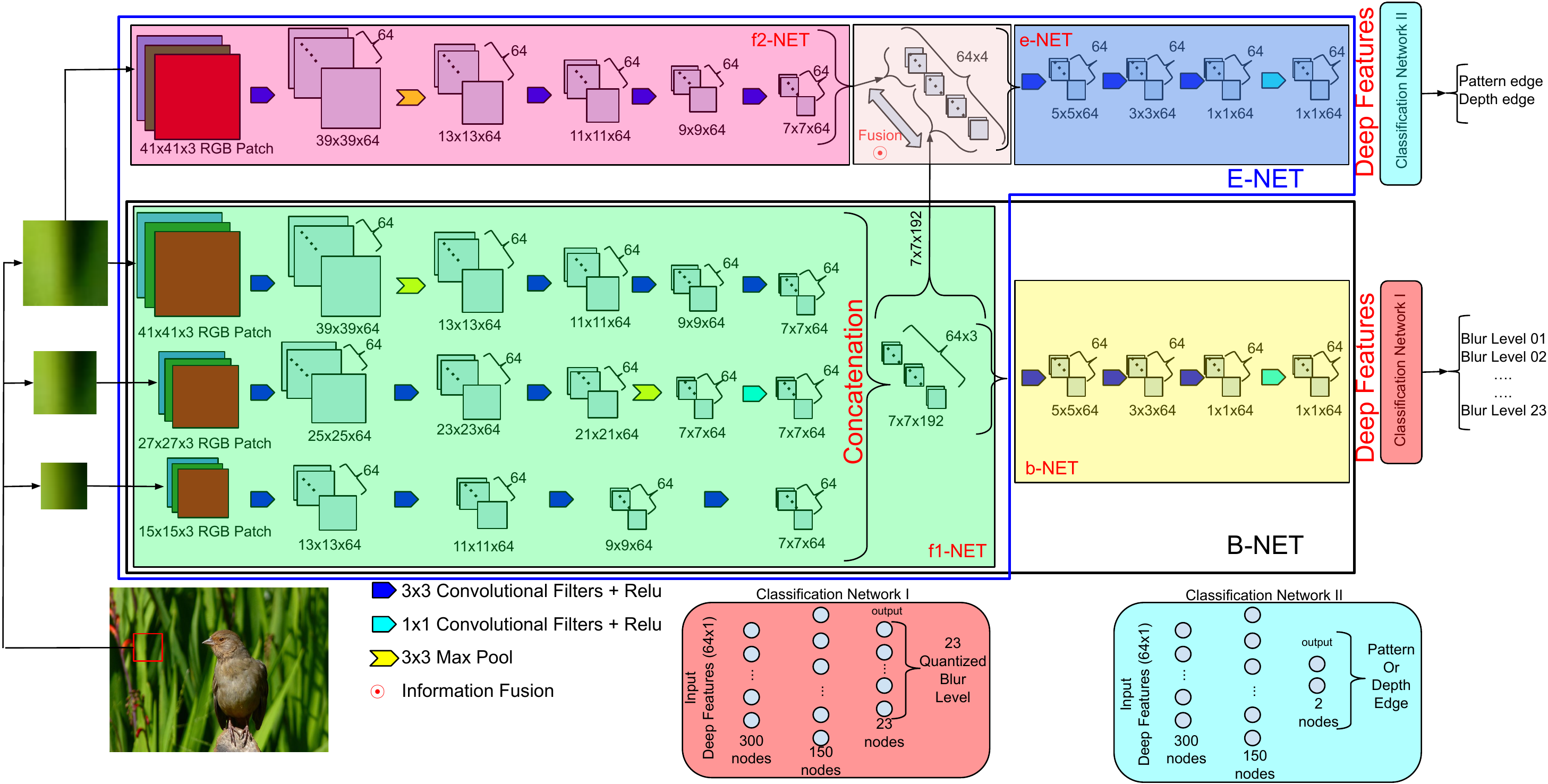}
        \vskip -0.125cm
        \caption{\label{fig:networkdesig} Overview of the feature extraction networks for \textit{depth} and \textit{pattern} edge separation and defocus blur estimation.}
    \end{center}
    \vskip -0.750cm

\end{figure*}

\subsection{Blur Estimation Network}

Our blur estimation network (\textbf{\textit{B-NET}}) is fed with image patches centered at \textit{pattern} edges. As noted in~\cite{patchsize1}, determining an appropriate patch size is a key issue in order to avoid patch scale ambiguities. In the context of blur estimation, Park et al.~\cite{park2017unified} used an edge strength measure to determine suitable patch sizes, assuming that strong edges are likely to be less blurry than weak edges. Claiming that blurrier patches require more spatial information in the representation, they used larger patches for weak edges and smaller patches for strong edges (since their network uses fixed-input patches, a padding procedure is adopted for small patches).
Despite the good results shown by the authors, we contend that edge strength does not necessarily correlate to blurriness. In fact,  a sharp low-contrast edge can be detected as a weak edge. In this work, we do not explicitly select a desired input patch size, but instead we propose a multi-scale model to fuse information at different resolution levels.

 \textbf{\textit{B-NET}} consists of two cascaded sub-networks: \textit{f1-NET} (green shaded box in Fig.~\ref{fig:networkdesig}) and \textit{b-NET} (yellow shaded box in Fig.~\ref{fig:networkdesig}). Sub-network \textit{f1-NET}  receives three patches of different sizes ($P^B_1 = 41 \times 41$, $P^B_2 = 27 \times 27$ and $P^B_3 = 15 \times 15$), which are extracted by centering the different-sized patches to the same edge location. These multi-scale patches, after a series of convolutional filters, Rectified Linear Units (ReLUs) and max pooling layers, are concatenated at the point where they reach the same spatial size (i.e. information at different resolution levels are fused), aiming to extract low-level blur information regarding the edges in a multi-scale way. It is important to note that we tried the same architecture using a single patch size (we tested different sizes), and the multi-scale approach gave better results than any individual patch size.

The output of \textit{f1-NET} is then fed to  \textit{b-NET}, which consists of another sequence of convolutional filters with ReLU activation functions. The goal of \textit{b-NET} is to extract deep features $f_B$ specialized to encode the blur level of the multi-scale patches. These  features $f_B \in  \mathbb{R} ^ {64 \times 1}$ are then sent to a classification network called ``Classification Network I'', which consists of three fully connected layers. This network classifies the input feature vector $f_B \in  \mathbb{R} ^ {64 \times 1}$ as one of the quantized blur levels ($23$ levels\footnote{Please check Section~\ref{subsec:dataprep} for details}) through two hidden layers ($300$ and $150$ nodes of each) and one output layer with $23$ nodes. ReLU activation functions are used in both hidden layers, and the \textit{softmax} classifier is used in the last layer.

\subsection{Depth vs. Pattern Edge Classification}

Typical edge-based defocus blur estimation methods~\cite{DefocusPaper,TIPpaper2012,icip2016fast,fixedsigma1} start by applying an edge detection scheme to the input image, and estimate a blur value for \textit{every} edge pixel. However, as discussed in Section~\ref{sec:intro}, edges related to a depth discontinuity do not present a well-defined blur value, since they are affected by more than one CoC. In this work, we  distinguish \textit{depth} from \textit{pattern} edges using  a deep CNN called \textbf{\textit{E-NET}}, with an architecture as illustrated in the top of Fig.~\ref{fig:networkdesig}.  

\textbf{\textit{E-NET}} is a network designed to extract local deep features to distinguish \textit{pattern} from \textit{depth} edge points.  Our hypothesis is that the low-level features extracted by \textit{f1-NET} using multi-scale patches encodes relevant information not only for blur estimation, but also for edge classification. As such, \textbf{\textit{E-NET}}  inherits all the weights from \textit{f1-NET}), and it also includes a separate branch that is fed with a fixed-sized patch ($P^E_1 = 41 \times 41$) that aims to extract features tailored to the edge classification problem. This branch is called sub-network \textit{f2-NET} and is shown in a purple shaded box in Fig.~\ref{fig:networkdesig}.

The outputs of \textit{f1-NET} (low-level blur information) and \textit{f2-NET} (edge classification features)
are then fused together  when they reach the same spatial resolution (see ``Information Fusion'' in Fig.~\ref{fig:networkdesig}). Fused information is then sent to \textit{e-NET}, which consists of a set of convolutional layers with ReLU activation functions to extract deep features $f_E$ specialized for \textit{pattern} and \textit{depth} edge classification. Finally, these  deep features $f_E \in  \mathbb{R} ^ {64 \times 1}$ are sent to a classification network called ``Classification Network II'', which consists of $2$ fully connected hidden layers of size $300$ and $150$ nodes each with ReLU activations functions,  similar to the the classification layers of \textbf{\textit{B-NET}}. However, the output layer presents only two nodes (\textit{pattern} or \textit{depth} edge) with  \textit{softmax} activation.

It is also important to note that there is significatly more labeled data for blur estimation than edge classification (in fact, we had to label data for this task). Hence, \textit{f1-NET} learns low-level blur features with a large amount of data, while  \textit{f2-NET} learns specific low-level information for edge classification with a more modest amount of data.



\subsection{Full Blur Map}

The output of \textbf{\textit{B-NET}}  is a sparse and quantized blur map computed only at \textit{pattern} edges provided by \textbf{\textit{E-NET}}.  As in most edge-based methods~\cite{DefocusPaper,karaalijungICIP2016,TIP20162,park2017unified,OurTIP2018}, a propagation scheme is used to obtain a full blur map.
For this task, the majority of edge-based defocus blur estimation methods~\cite{DefocusPaper,karaalijungICIP2016,TIP20162,park2017unified} adopted the Laplacian-based colorization scheme~\cite{LaplacianM}.
However, this propagation scheme is time-consuming, and it respects the edges of the original blurry image $I^B$ even
at \textit{pattern} edges, causing visible artifacts at the full blur map even when depth (and blur) does not vary. 

An alternative approach would be to use a cross-domain filter that acts on both the sparse blur map $I_S$ and the binary edge map $E$, assuming that $I_S(\bm{x}) = 0$ when $E(\bm{x}) = 0$. If function $\beta$ defines an adaptive weight for each pair of pixels $(\bm{x}, \bm{y})$, the interpolated blur map $B$ is given by 
\begin{equation}
B(\bm{x}) = \frac{\sum_{\bm{y}\in \mathcal{N}_{\bm{x}}} \beta(\bm{x},\bm{y})I_S(\bm{y})}
{\sum_{\bm{y}\in \mathcal{N}_{\bm{x}}} \beta(\bm{x},\bm{y})E(\bm{y})} = 
\frac{\mathcal{F}(I_S, I)}
{\mathcal{F}(E, I)},
\label{eq:interp}
\end{equation}
where $\mathcal{N}_{\bm{x}}$ is the neighborhood around pixel $\bm{x}$ that specifies the data propagation region (and can be the whole image), and $\mathcal{F}(J,I)$ denotes the cross-domain  filtering of image $J$ using image $I$ as the guidance (the selection of the filtering approach impacts the weights $\beta$). The normalization (denominator) ensures that the contribution of all valid values within the neighborhood $\mathcal{N}_{\bm{x}}$
adds up to one. 

Edge-aware filters (such as the bilateral filter) consist of adaptive weights $\beta$ that try to prevent mixing information across image edges. In our problem, however, we can (and should) allow propagation across \textit{pattern} edges, but want to prevent propagation across \textit{depth} edges, since they typically separate objects at different depths (and hence, blur values).
One edge-aware filter that allows a suitable adaptation for treating \textit{pattern} and \textit{depth} edges in a different way is the 
Domain Transform (DT) filter presented in~\cite{Gastal}. The core idea of the DT filter is to perform 1D domain transformations such that samples considered ``different'' are spatially ``pushed apart'' from each other. With this transformation, a convolution kernel with a fixed size acts like an edge-preserving filter (similarly to the bilateral filter), and two-dimensional filtering is achieved by alternating filtering along rows and columns. For more details on the DT filter, the reader is directed to~\cite{Gastal}.

In this work, we propose a simple modification of the domain transformation that adds a penalty to \textit{depth} edges.
The proposed domain transformation function $ct$ (in the continuous domain) that measures the distance between two 1D points $u \leq w$ is given by
\begin{equation}
    ct(u,w) = \int_{u}^{w} 1 + \Psi(x) + \frac{\sigma_s}{\sigma_r} \sum_{k=1}^{c} | I'_k(x) |  dx,
    \label{eq:domain_transform_modified}
\end{equation}
where $I'_k$ denotes the derivative of the $k$-th color channel of guide image $I$ (along a line or column), and $\sigma_s$,  $\sigma_r$ are parameters that control the spatial and color range of the kernel, as defined in~\cite{Gastal}. Note that the value of $ct(u,w)$ progressively increases as does the spatial and color distances between points $u$ and $v$. Our modification to~\cite{Gastal} is the introduction of the cost term
$\Psi(x)$ given by
\begin{equation}
\Psi(x) = \left\{
\begin{array}{ll}
\psi, & \text{~if~} x \text{~is a depth edge}\\
0, & \text{otherwise} 
\end{array}
\right .,
\label{eq:costval}
\end{equation}
and $\psi$ is a constant that defines the penalty introduced by depth edges. Note that $1 + \psi$ becomes a lower bound for the distance between any to points $u$ and $w$ separated by a \textit{depth} edge. Hence, if $\psi$ is sufficiently large, blur propagation between these two points is virtually stopped. The domain-transformed signal is then convolved with a low-pass filter with variance $\sigma_s^2$ (very fast implementation if box filters are used).

Since fine textures and/or noise can block blur propagation across \textit{pattern} edges, the input blurry image $I^B$ is first simplified by filtering with the edge-aware filter itself (i.e., $\hat{I^B} = \mathcal{F}(I^B, I^B)$), then is used as the guidance image in the propagation given by Eq.~\eqref{eq:domain_transform_modified}. Note that a similar strategy was employed in~\cite{OurTIP2018}, but without considering \textit{depth} edges.


In terms of complexity, \textbf{\textit{B-NET}} and \textbf{\textit{E-NET}} require, respectively, 123.46 and 149.16 MFLOPs. Since these networks are fed with edge-centered patches, the main bottleneck of the proposed method relates to the total number of edges. The interpolation step is very fast, since the DT-filter can be implemented in linear time~\cite{Gastal}.

\section{Experiments}
\label{sec:experiments}

\subsection{Data Preparation}
\label{subsec:dataprep}

We use images from then ILSVRC~\cite{ILSVRC15}, MS-COCO~\cite{MSCOCO} and HKU-IS~\cite{saliencyData} datasets in order to train the proposed network architectures. Due to the lack of annotated datasets with blur data, we strongly rely on synthetic data.

\textbf{Synthetic data for defocus blur estimation:} in order to generate a dataset with known ground truth blur value (i.e. defocus blur amount is know for a given pixel) to train \textit{\textbf{B-NET}}, we first manually select sharp (all-in-focus) images that do not contain any visually detectable blurry pixels (250 from ILSVRC and 250 from MS-COCO), and then convolve each selected sharp image $I^S$ with a blur kernel. {Although the actual kernel depends on the camera and lens system, a disk kernel models a perfect lens system with a circular aperture, and was shown to be a good approximation in experiments with real images conducted in~\cite{ZhuFREQ}. Here, we used disk kernels $C_{R_{i}}$ to generate blurry images $I^B$ through
\begin{equation}
\label{eq:blurringtheimage}
    I^B_i=I^S * C_{R_{i}},~ i \in \{1,2,\cdots, S\},
\end{equation}
where $R_i$ denotes the blur level (i.e., the kernel size), and $S$ is the number of quantized blur values. 
Following~\cite{TIPpaper20161}, we set $S=23$, starting from \ali{$R_1 = 0.5$} and increasing up to $R_{23} = 6$ with a step size of $0.25$. 
As noted in~\cite{liu2020defocus}, the chosen range contemplates the expected blur amount in most image resolutions, except for extremely defocused regions in ultra-high resolution images. Then, we extract approximately 5M patches from the edge points of synthetically blurred images, assuring that they are equally distributed for each blur level.  Although blurring the whole image with a single spatially-invariant blur kernel is clearly an oversimplification since it does not impose any blur variations due to depth changes, this approach has generalized well to the blur estimation problem, especially in region-based methods~\cite{CVPR2010FREQ,ZhuFREQ}. Furthermore, the separation of \textit{depth} and \textit{pattern} edges from each other assures that
the proposed network will only be fed by patches that do not present strong depth variations.

\begin{figure}[!h]
    \begin{center}
        \includegraphics[width=0.3\columnwidth]{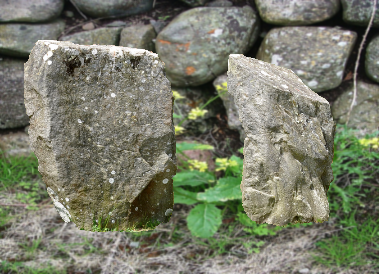}
        \includegraphics[width=0.3\columnwidth]{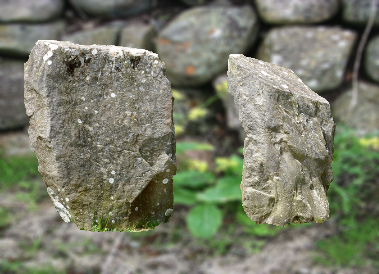}
        \includegraphics[width=0.3\columnwidth]{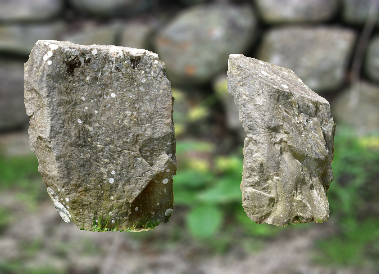} \\
        \includegraphics[width=0.3\columnwidth]{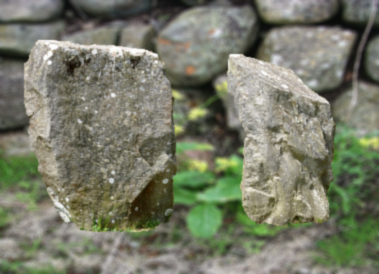}
        \includegraphics[width=0.3\columnwidth]{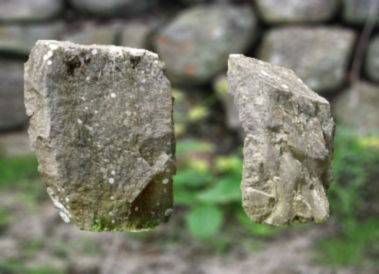}
        \includegraphics[width=0.3\columnwidth]{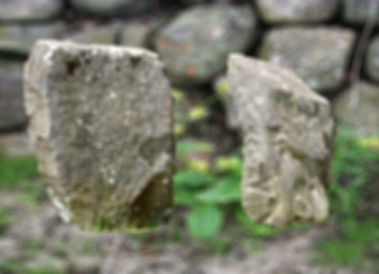}
        \caption{\label{fig:synforback} \ali{Synthesized foreground-background images. Foreground blur level $R_{k_1}$, background blur level $R_{k_2}$. First row: $R_{k_1}=0$ and  $R_{k_2}=1,3,5$, from left to right. Second row: $R_{k_1}=1$ and $R_{k_2}=3$, $R_{k_1}=1$ and $R_{k_2}=5$, and $R_{k_1}=3$, $R_{k_2}=5$ from left to right. }}
    \end{center}
     \vskip -0.750cm
\end{figure}

\textbf{Synthetic data for depth vs. pattern edge separation:} to train \textit{\textbf{E-NET}}, which distinguishes \textit{depth} edges from \textit{pattern} edges, we need edge points that present different blurriness (i.e. depth) levels at different sides. Although a database with similar characteristics is reported in~\cite{understandblur}, it was not made publicly available. Due to the lack of annotated data, we produced synthetic scenes $I^{FB}$ in a foreground-background manner using 200 salient regions as foreground objects from the HKU-IS~\cite{saliencyData} dataset, which provides images $S^I$ with binary masks $S^B$ indicating salient objects. We also use 100 images from ILSVRC~\cite{ILSVRC15} and 100 images from MS-COCO~\cite{MSCOCO} datasets to compose the background of our dataset.

To generate our synthetic dataset, we first crop the salient object from the salient image $S^I$ using the provided binary mask $S^B$. Then, we blur the cropped region and its corresponding mask image with a disk blur kernel \ali{$C_{R_{k_1}}$}, while simultaneously blurring a sharp (focused) image with a different disk blur kernel \ali{$C_{R_{k_2}}$} (with \ali{$R_{k_1} < R_{k_2}$}), which will be the background image. Finally, we alpha-blend the blurry cropped image onto the blurry background image using the blurred binary mask as alpha values. We used four levels of blur scale for this task:
\ali{$R_{1}=0$} for no blur, \ali{$R_{2}=1$}  for low blur, \ali{$R_{3}=3$} is for medium blur and \ali{$R_{4}=5$} is for high blur.

For each salient image -- background image pair, we synthesize $N=6$ synthetic foreground-background images $I^{FB}_n, n \in \{1, 2, \ldots,N\}$, \ali{as illustrated in Fig.~\ref{fig:synforback}}. 
Finally, we extract around 2M \textit{depth} edge patches from the boundary of projected objects. Since using only synthetic images could overfit the network, we also manually labeled \textit{depth} edges in real images, \ali{as illustrated in Fig.~\ref{fig:realdata}.} 
More precisely, we labeled 100 real defocused images chosen from Flicker, from which 500K \textit{depth} edge patches were extracted. We also perform data augmentation by randomly rotating the images by $-90$, $-30$ $60$, $135$, or $180$  degrees. 
Since noise did not seem to be an issue with the blurry images in our datasets, we did not add noise in the augmentation process.   

\begin{figure}[b]

    \begin{center}
        \includegraphics[width=0.45\columnwidth]{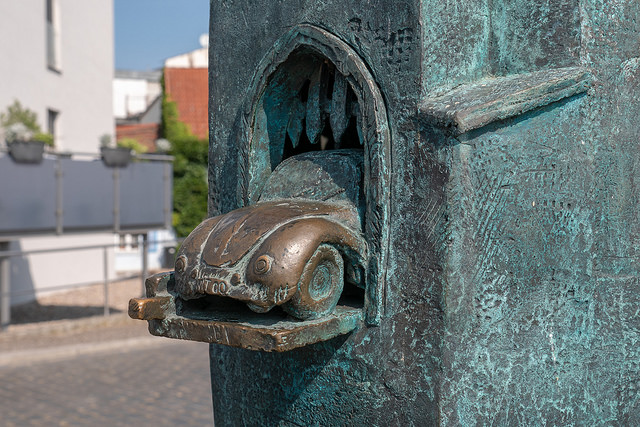}
        \includegraphics[width=0.45\columnwidth]{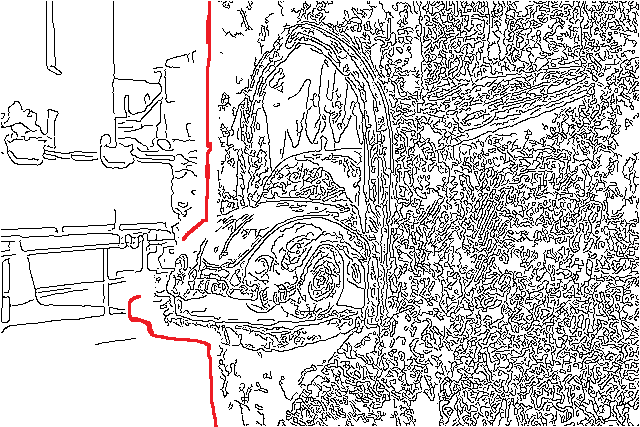} \\
        \caption{\label{fig:realdata} Manually labeled depth edges from a defocused image in Flickr~\cite{flick_photo}}
   \end{center}
   
\end{figure}

As for \textit{pattern} edge patches, which is the other class label of \textit{\textbf{E-NET}}, we extract approximately 2.5M patches from 100 sharp (focused) images of ILSVRC~\cite{ILSVRC15} and 100 sharp (focused) images of MS-COCO~\cite{MSCOCO} datasets, synthetically blurring them with different blur patterns such as uniform blur, gradually changing blur, and step-wise changing blur (with step size $0.25$ to simulate small blur changes at the same depth layers).

\subsection{Model Training}

As described in Section~\ref{sec:proposed}, although there is some weight sharing between \textbf{\textit{B-NET}} and \textbf{\textit{E-NET}}, the two networks are trained separately. Both models were implemented using Tensorflow with $80-20\%$ train-validation data separation, with the \textit{softmax cross entropy} as the loss function. We used the Adam~\cite{adamop} optimizer with a batch size of $256$.

We start by first training \textbf{\textit{B-NET}}, using an initial learning rate of $10^{-3}$ that is divided by $10$ every $10$ epochs, yielding convergence after $75$ epochs. In a subsequent step we trained \textbf{\textit{E-NET}} with the same initial learning rate, but divided by $10$ at every $5$ epochs instead of $10$. It is important to recall that during the training of \textbf{\textit{E-NET}}, the shared weights from \textbf{\textit{B-NET}} (called \textit{f1-NET}, as shown in Fig.~\ref{fig:networkdesig}) are frozen, and the remaining weights of \textbf{\textit{E-NET}} are learned. With this strategy, the network converged after $30$ epochs.

\begin{figure}[!t]

    \begin{center}
        \includegraphics[width=0.3\columnwidth]{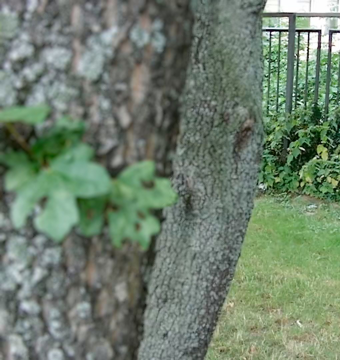} 
        \includegraphics[width=0.3\columnwidth]{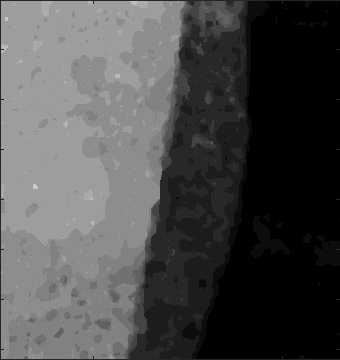}
        \includegraphics[width=0.3\columnwidth]{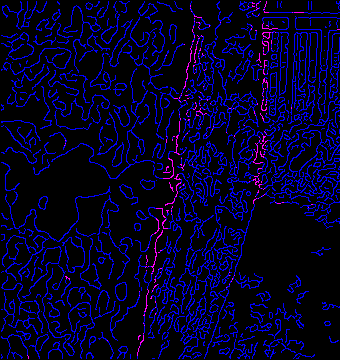} \\
        \includegraphics[width=0.3\columnwidth]{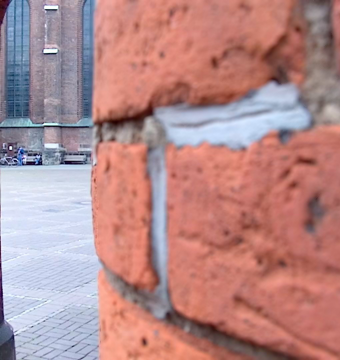}
        \includegraphics[width=0.3\columnwidth]{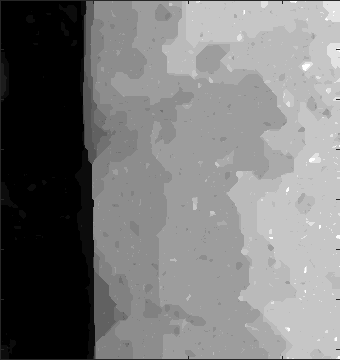}  
        \includegraphics[width=0.3\columnwidth]{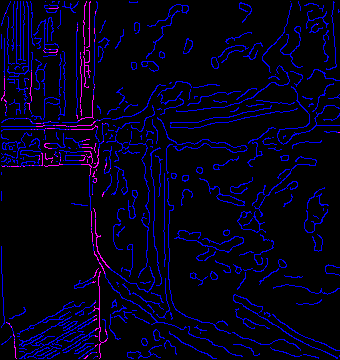} \\
        \caption{\label{fig:resultEdge} \textit{Depth} and \textit{pattern} edge separation via \textbf{\textit{E-NET}}. From left to right (in pairs): original images (from ~\cite{TIPpaper20161}), ground truth blur map, and corresponding \textit{depth} edges (red) and \textit{pattern} edges (blue).}
    \end{center}
    \vskip -0.75cm
\end{figure}

\textbf{Implementation details:} Regarding defocus blur interpolation, we used $\sigma_{r_1} = 0.5$ and $\sigma_{s_1} = 7$ to get the simplified image $\hat{I^B}$, and $\sigma_{r_2} = 3.75$ and $\sigma_{s_2} = \min\{H,W\}/8$ to propagate sparse blur estimates. $H$ and $W$ are the height and weight of the simplified defocus image $\hat{I^B}$ respectively. Before the simplification operation, the image pixels are normalized to the range $[0, 1]$. Since the data interpolation has to be propagated to the whole image from edge pixels (which might be very sparse), we chose a large spatial kernel size $\sigma_{s_2}$. The range kernel $\sigma_{r_2}$ is also a relatively large value, chosen to increase pixel similarity and allow some propagation over \textit{pattern} edges. For the penalization term of \textit{depth} edges, we empirically set $\psi = 100$ (which blocks propagation almost entirely) and used this default value in all experiments -- larger values for $\psi$ show almost no difference from the default value.

 \begin{table*}[!ht]
\caption{Quantitative evaluation of blur maps for the dataset provided in~\cite{TIPpaper20161} in terms of Average MAE $\pm$ Standard Deviation (STD) \ali{of Raw Blur},  \ali{Average MAE of Relative Blur}, Running Time and Implementation.}
\centering
\scalebox{0.95}{
\begin{tabular} {c c c c c c c c} 
 \hline
  &~\cite{DefocusPaper}&~\cite{TIPpaper20161}&~\cite{OurTIP2018}&~\cite{park2017unified}&~\cite{Lee_2019_CVPR} &~\cite{liu2020defocus} &Our\\ [0.5ex] 
 \hline\hline
MAE $\pm$ STD \ali{of Raw Blur} & $0.629\pm 0.205$ & $0.197\pm 0.072$ & $0.378\pm 0.170$ & $0.307\pm 0.121$ & $0.611\pm 0.217$ & $0.180 \pm \mathbf{0.067}$ & $\mathbf{0.169}\pm 0.068$ \\

\ali{MAE of Relative Blur} & 0.098  &  0.033  &  0.063  &  0.049 &   0.101 &   0.030  &  \textbf{0.028} \\
Time (sec.) & $5.60$ & $\thicksim 114$ & $0.69$ & $7.02$  & $\mathbf{0.14}$  &  $\thicksim 184^*$ & $4.55$ \\ 

\hline
Impl.        & Matlab    & C++ \& Matlab         & Matlab    & Matlab   & Python/GPU  & Not informed   & Python/GPU \\ 
 \hline
 \multicolumn{8}{l}{$^*$ Running time extracted from \cite{liu2020defocus}, where a different hardware was used. Authors report running time higher than~\cite{TIPpaper20161}.}
\end{tabular}
}
\label{table:1}
    \vskip -0.25cm
\end{table*}

\subsection{Experimental Validation}

In order to validate the proposed defocus blur estimation method, we use the dataset introduced in~\cite{TIPpaper20161} that contains $22$ real defocused images with resolution $360 \times 360$ captured by a Light Field camera, and it is (to our knowledge) the only dataset in the literature that provides the ground truth defocus blur values for each pixel. Although it also presents images corrupted by artificial noise, we restricted our analysis to the subset with natural image noise only since our model was not trained with artificial noise. We use the popular Mean Absolute Error (MAE) to quantitatively compare our defocus blur estimation approach with SOTA methods, and show edge map images with highlighted \textit{depth} edges for qualitative evaluation of \textbf{\textit{E-NET}} on the same dataset.

\begin{figure*}[t]
    \begin{center}

        \includegraphics[width=0.21\columnwidth]{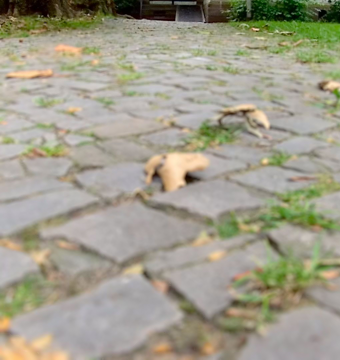}
        \includegraphics[width=0.21\columnwidth]{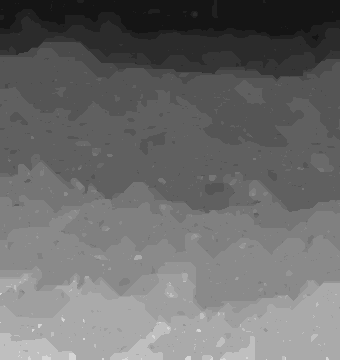} 
        \includegraphics[width=0.21\columnwidth]{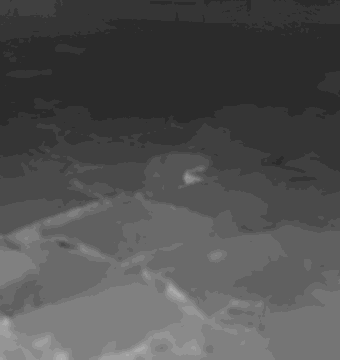} 
        \includegraphics[width=0.21\columnwidth]{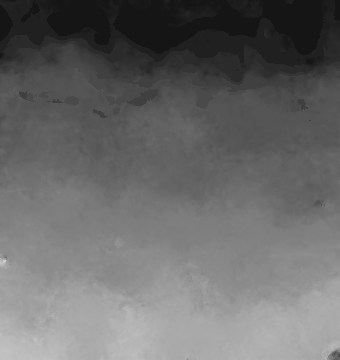} 
        \includegraphics[width=0.21\columnwidth]{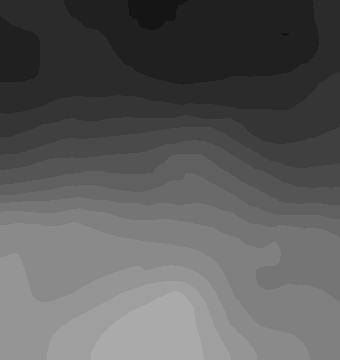} 
        \includegraphics[width=0.21\columnwidth]{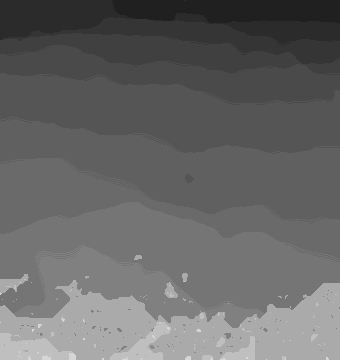} 
        \includegraphics[width=0.21\columnwidth]{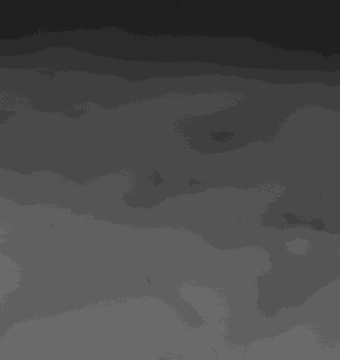}
        \includegraphics[width=0.21\columnwidth]{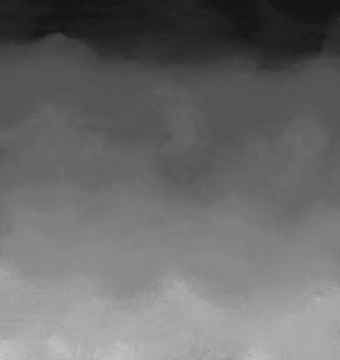}
        \includegraphics[width=0.21\columnwidth]{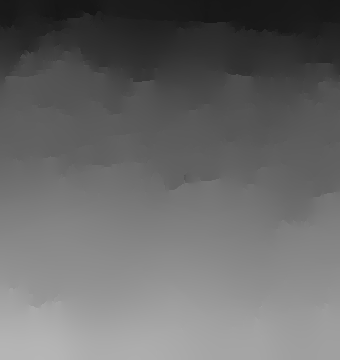} \\
        \includegraphics[width=0.21\columnwidth]{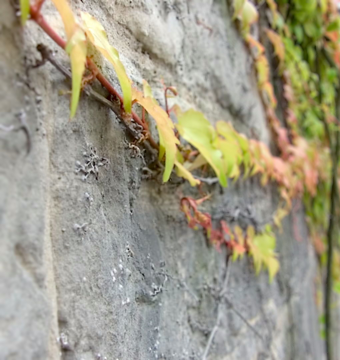}
        \includegraphics[width=0.21\columnwidth]{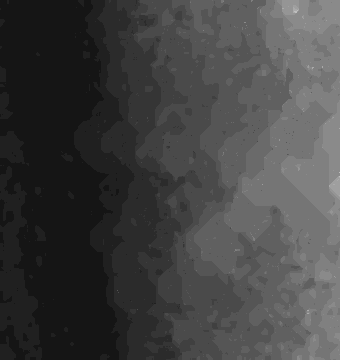} 
        \includegraphics[width=0.21\columnwidth]{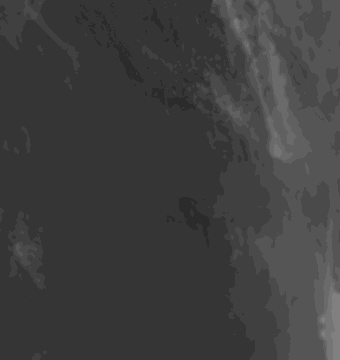} 
        \includegraphics[width=0.21\columnwidth]{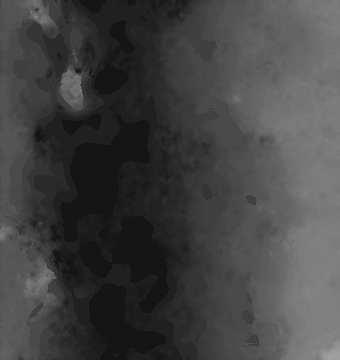} 
        \includegraphics[width=0.21\columnwidth]{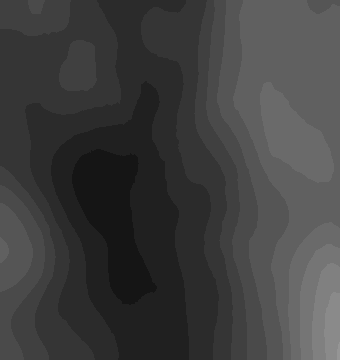} 
        \includegraphics[width=0.21\columnwidth]{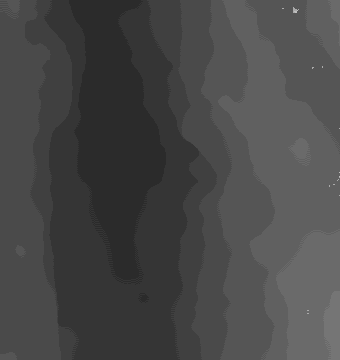} 
        \includegraphics[width=0.21\columnwidth]{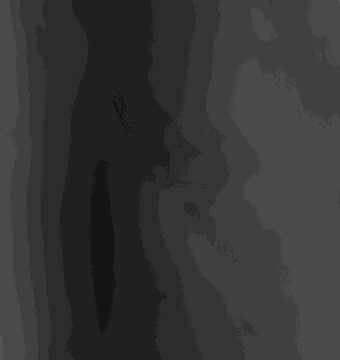}
        \includegraphics[width=0.21\columnwidth]{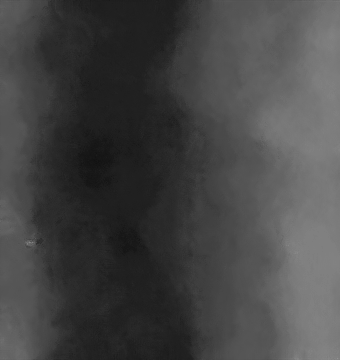}
        \includegraphics[width=0.21\columnwidth]{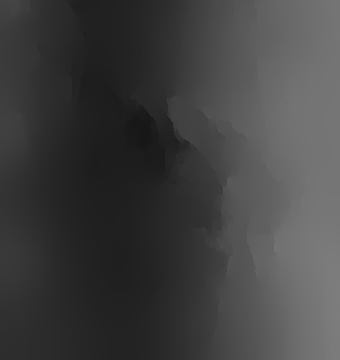} \\
        \includegraphics[width=0.21\columnwidth]{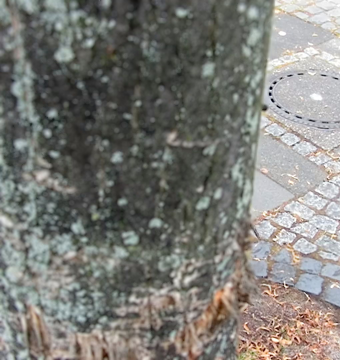}
        \includegraphics[width=0.21\columnwidth]{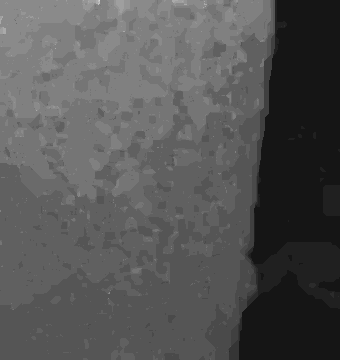} 
        \includegraphics[width=0.21\columnwidth]{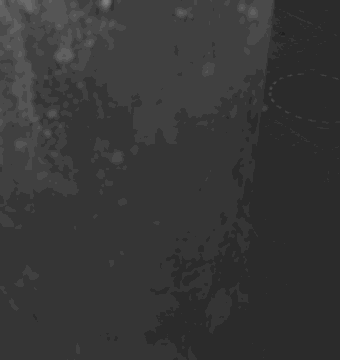} 
        \includegraphics[width=0.21\columnwidth]{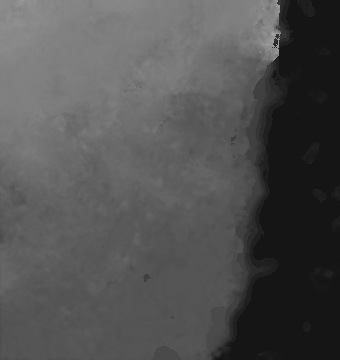} 
        \includegraphics[width=0.21\columnwidth]{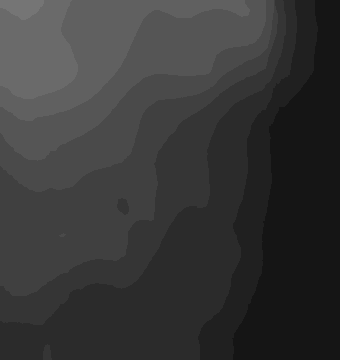} 
        \includegraphics[width=0.21\columnwidth]{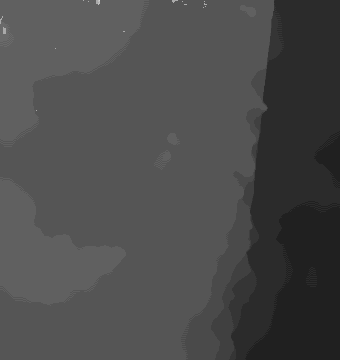} 
        \includegraphics[width=0.21\columnwidth]{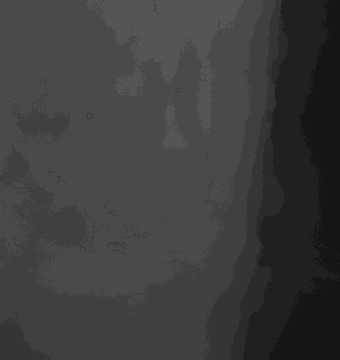}
        \includegraphics[width=0.21\columnwidth]{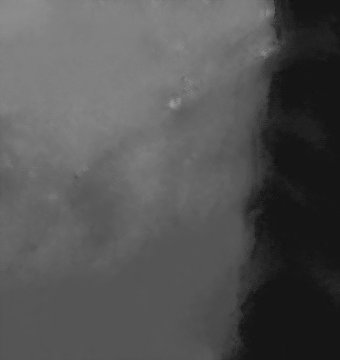}
        \includegraphics[width=0.21\columnwidth]{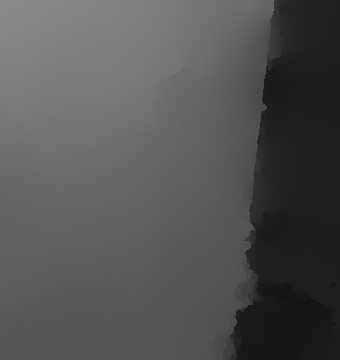} \\
        \includegraphics[width=0.21\columnwidth]{orig_22.png}
        \includegraphics[width=0.21\columnwidth]{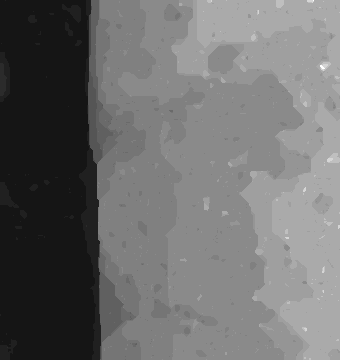} 
        \includegraphics[width=0.21\columnwidth]{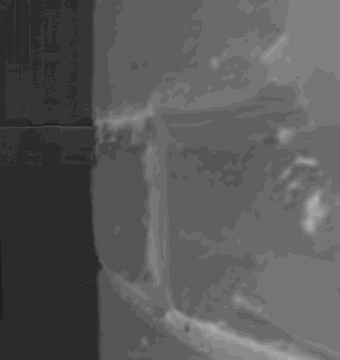} 
        \includegraphics[width=0.21\columnwidth]{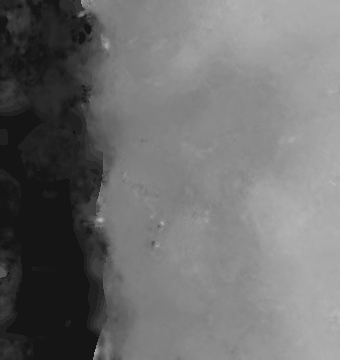} 
        \includegraphics[width=0.21\columnwidth]{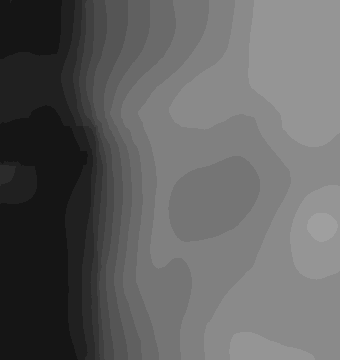} 
        \includegraphics[width=0.21\columnwidth]{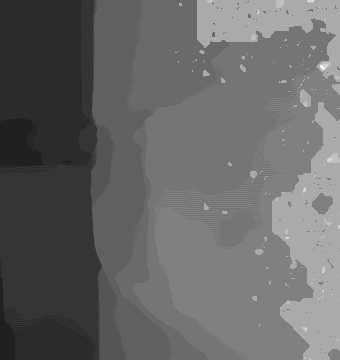} 
        \includegraphics[width=0.21\columnwidth]{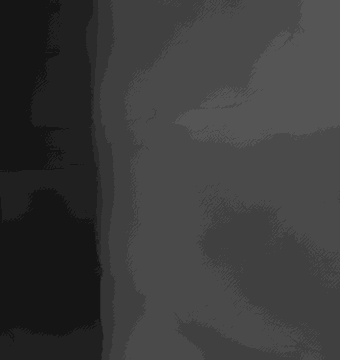}
        \includegraphics[width=0.21\columnwidth]{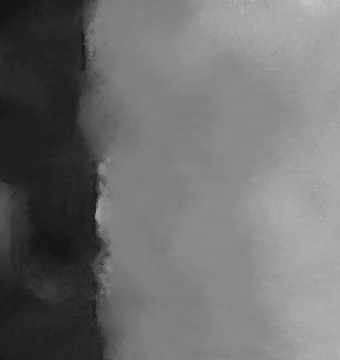}
        \includegraphics[width=0.21\columnwidth]{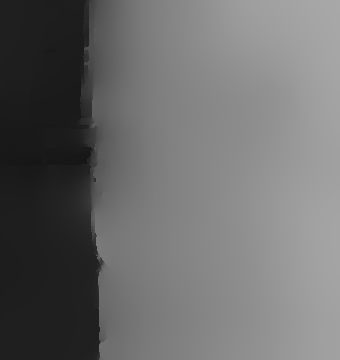} \\
        \caption{\label{fig:resultOnTipDataOrig} Comparison of blur estimation algorithms using the dataset provided in~\cite{TIPpaper20161}. From left to right: original image, ground truth,  results of~\cite{DefocusPaper},~\cite{TIPpaper20161},~\cite{OurTIP2018},~\cite{park2017unified},~\cite{Lee_2019_CVPR},~\cite{liu2020defocus} and the proposed method. }
    \end{center}
     \vskip -0.75cm
\end{figure*}

We first illustrate examples of \textit{pattern} vs. \textit{depth} edge classification produced by \textbf{\textit{E-NET}} in Fig.~\ref{fig:resultEdge}. 
In the first blurry image, three different depth layers can be seen (from left to right), and \textbf{\textit{E-NET}} manages to distinguish most of the edge points that present depth discontinuities (abrupt blur change). 
In the second image, there is an abrupt depth transition from the red wall to the background, and \textbf{\textit{E-NET}} correctly labels these boundary points as \textit{depth} edges, with a few false negatives. Note that edge classification is an intermediate step of our approach, and it will be evaluated implicitly by showing that it does improve the final
goal (dense blur estimation), as will be shown next. Furthermore, the amount of data used to train \textbf{\textit{E-NET}} is rather limited, so we only split the data into train and validation (no test set). For the sake of illustration, the accuracy in the validation set was 88\%.


We computed the MAE \ali{of the raw blur values} for each of the $22$ images in the database and reported the average MAE and the standard deviation of the full blur maps obtained by the proposed method\footnote{Our code is available at \url{github.com/alikaraali/DepthEdgeAwareBENet}} and competitive approaches in Table~\ref{table:1}. \ali{We also computed and reported the MAE of the relative blur values by re-scaling the raw blur values to $[0,1]$, as done in~\cite{Lee_2019_CVPR}}. We can see that our method outperforms all the competitive approaches \ali{in both raw and relative blur}\footnote{\ali{Although we used official implementation of~\cite{Lee_2019_CVPR} from \url{github.com/ake/DMENet}, we obtained an average relative blur MAE slightly different from the value reported in their paper.}}. It is slightly better than region-based methods~\cite{TIPpaper20161,liu2020defocus}, which are considerably slower. Please note that as the method ~\cite{Lee_2019_CVPR} produces a ``relative blur map'', they normalize the GT blur map by the maximum value to report their accuracy. For a fair comparison with the other approaches, we do the opposite: multiply their blur map by the maximum blur value to report the raw blur accuracy. Since the methods described in~\cite{DefocusPaper,OurTIP2018,park2017unified} model the blur with a Gaussian PSF, we re-scale their results from Gaussian scale to disk radii via the mapping function provided by~\cite{TIPpaper20161}. Also, since the method presented in~\cite{park2017unified} is trained with a maximum Gaussian blur scale $\sigma_g=2.0$, we clipped the ground truth values to $2$ when computing the results of~\cite{park2017unified},  which favored their results significantly. The average running times for all the analyzed methods considering all the images are also shown in Table~\ref{table:1}. Although~\cite{Lee_2019_CVPR} has the fastest execution speed, our method presents a very good compromise between MAE and running time when compared to other SOTA methods. The supplementary material provides a comprehensive study of the dataset.

Since an important application related to blur estimation is deblurring, we have also evaluated how well our method integrates in this task. 
Following~\cite{TIPpaper20161}, we used the combination of the methods proposed in \cite{laplacianpriors} and \cite{levin2007image} as a deblurring baseline approach, 
and used as input the ground truth data provided in~\cite{TIPpaper20161}, the defocus maps produced by SOTA approaches and by the proposed method.
The average PSNR and SSIM values are summarized in Table~\ref{table:deblur}, showing that our method produced the highest average gain for both PSNR and SSIM, being inferior only to the deblurring results obtained with GT blur estimates. 
Fig.~\ref{fig:deblur} shows a visual comparison of some deblurring results, highlighting regions with high structural
or textural information. Our method presents visual results similar to those obtained with the ground truth blur map and the blur map produced with~\cite{liu2020defocus}, while being much faster than~\cite{liu2020defocus}. A detailed analysis of this experiment, along with the visual results of all methods, is provided in the supplementary material. 


\begin{table*}[ht]
\caption{Average PNSR \& SSIM for the deblurred images on the dataset provided in~\cite{TIPpaper20161} using different blur maps. Best value shown in bold, second best in italic.}
\centering
\scalebox{0.99}{
\begin{tabular} {c c c c c c c c c c} 
 \hline
   & Blurry Im. & GT Data &~\cite{DefocusPaper}&~\cite{TIPpaper20161}&~\cite{OurTIP2018}&~\cite{park2017unified}&~\cite{Lee_2019_CVPR} &~\cite{liu2020defocus} &Our\\ [0.5ex] 
 \hline\hline
 PSNR/SSIM & 24.20/0.782 &   \textbf{27.69}/\textbf{0.885} &   23.96/0.786 &   26.45/0.862 &   25.72/0.833 &   25.41/0.839 &   24.70/0.801 &  26.64/0.866 &  \textit{27.00}/\textit{0.870} \\
Gain       & N/A         & \textbf{3.48}/\textbf{0.102}    & -0.24/0.004   & 2.25/0.080    & 1.51/0.050    & 1.20/0.056    &  0.49/0.018.  &  2.44/0.084  & \textit{2.80}/\textit{0.088} \\
 \hline
\end{tabular}
}
\label{table:deblur}
\end{table*}

\begin{figure*}[!t]
    \begin{center}
        \includegraphics[width=0.31\columnwidth]{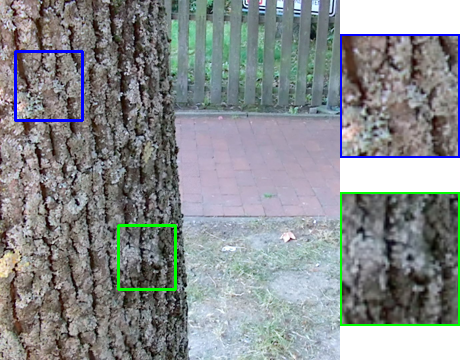}
        \includegraphics[width=0.31\columnwidth]{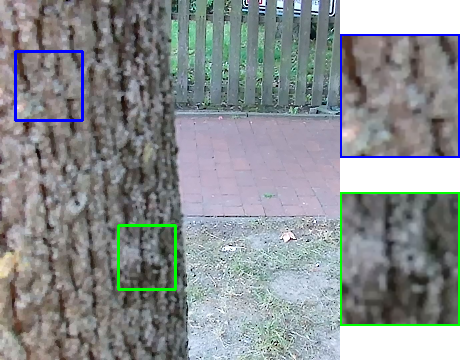}
        \includegraphics[width=0.31\columnwidth]{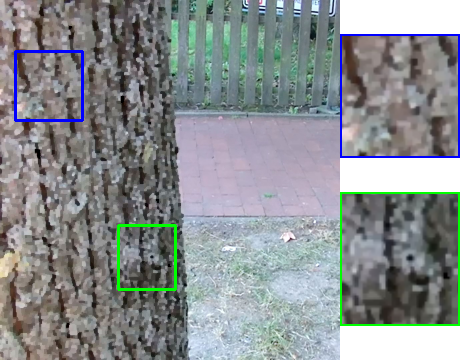}
        \includegraphics[width=0.31\columnwidth]{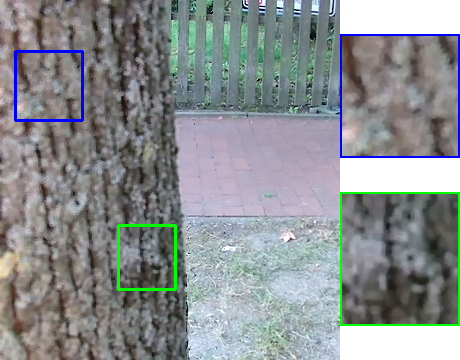}
        \includegraphics[width=0.31\columnwidth]{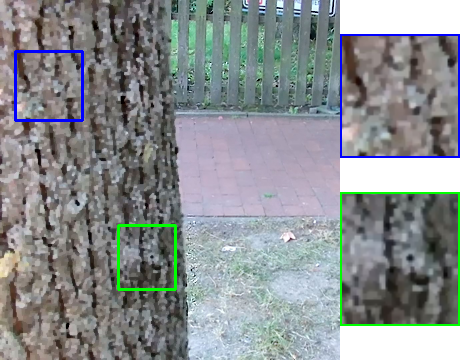}
        \includegraphics[width=0.31\columnwidth]{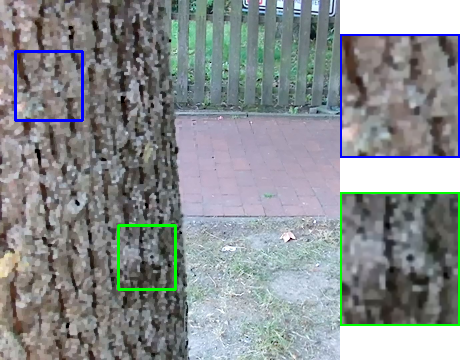}	\\			
				\vskip .01cm
        \includegraphics[width=0.31\columnwidth]{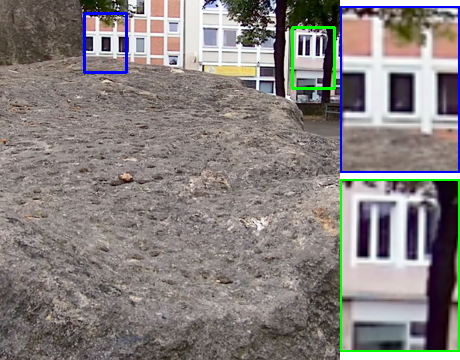}
        \includegraphics[width=0.31\columnwidth]{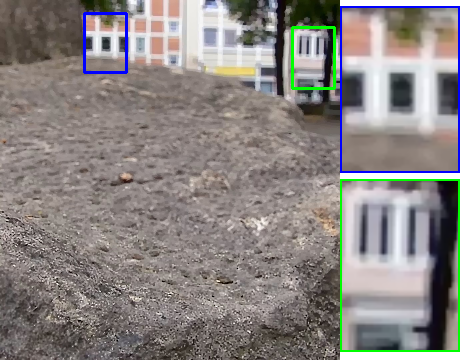}
        \includegraphics[width=0.31\columnwidth]{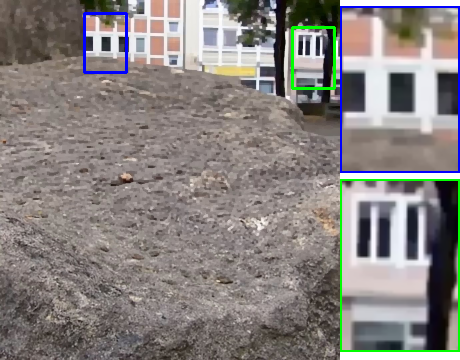}
        \includegraphics[width=0.31\columnwidth]{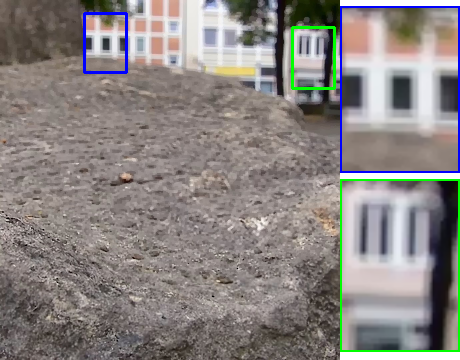}
        \includegraphics[width=0.31\columnwidth]{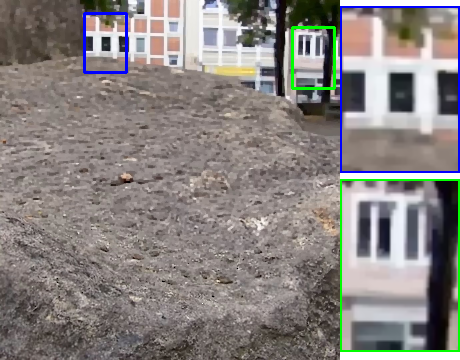}
        \includegraphics[width=0.31\columnwidth]{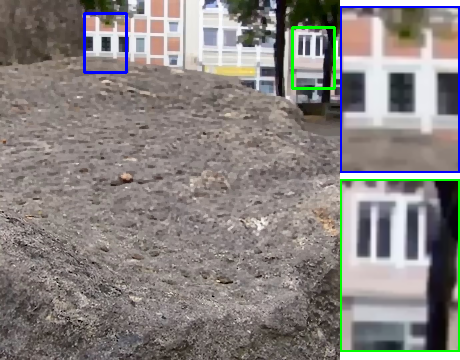}	\\			
				\begin{tabular}{UUUUUUU}
				(a) & (b) & (c) & (d) & (e) & (f)
				\end{tabular}
        \caption{\label{fig:deblur} Deblurring results using different blur maps as input. (a)-(b) sharp and blurry images. (c)-(f): results using the GT blur map,~\cite{Lee_2019_CVPR},~\cite{liu2020defocus} and the proposed method. }
    \end{center}
    \vskip -0.75cm
\end{figure*}

\begin{figure}[!hb]
\vskip -0.25cm
    \begin{center}
        \includegraphics[width=1.0\columnwidth]{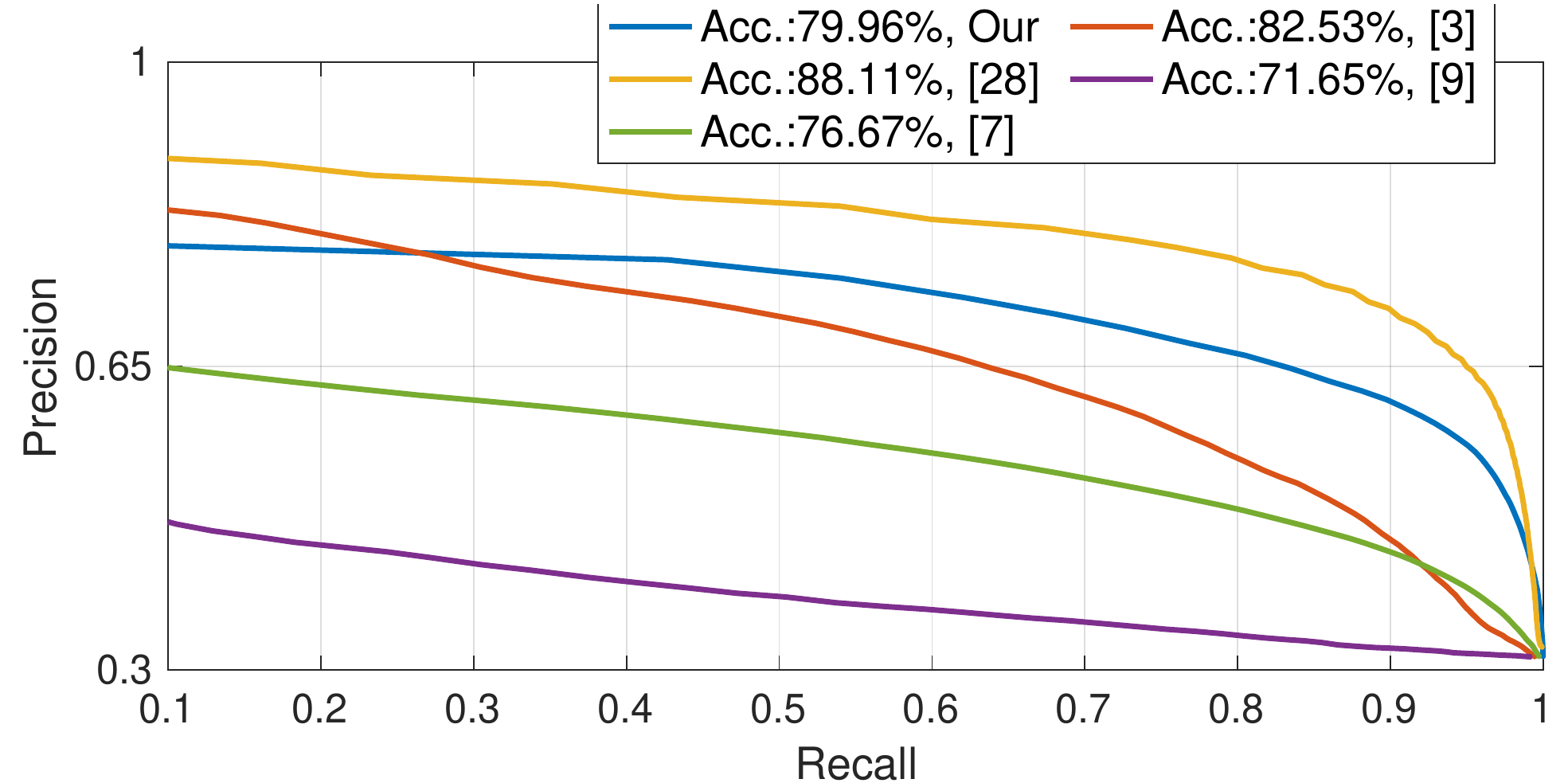}
        \caption{\label{fig:prcurve} Precision-recall curves and best accuracy for \cite{DefocusPaper}, \cite{OurTIP2018}, \cite{park2017unified}, \cite{Lee_2019_CVPR} and the proposed method on the CUHK defocus detection dataset \cite{cuhkdata}.}
    \end{center}
    \vskip -0.5cm
\end{figure}

Finally, we tested our method in the related task of defocus blur detection (DBD), which aims to find in-focus and out-of-focus regions of a given image. For this task, we use the CUHK blur detection dataset~\cite{cuhkdata}, which contains 704 defocused images along with the corresponding binary blur maps as ground truth. Since the main focus of the proposed method is defocus blur estimation (i.e., finding the actual blur level at each pixel), DBD is performed by thresholding the estimated blur maps. 
More precisely, we used the same adaptive thresholding approach adopted in~\cite{park2017unified} and~\cite{Lee_2019_CVPR}, which consists of defining a threshold 
\begin{equation}
\tau = \alpha v_{max} + (1 - \alpha) v_{min},
\end{equation}
where $v_{max}$ and $v_{min}$ are the maximum and the minimum blur values of the estimated defocus blur map, and $\alpha$ is an empirically chosen parameter, which is set so as to maximize the accuracy of each method individually.

\begin{figure*}[t]
    \begin{center}
        \includegraphics[width=0.32\columnwidth]{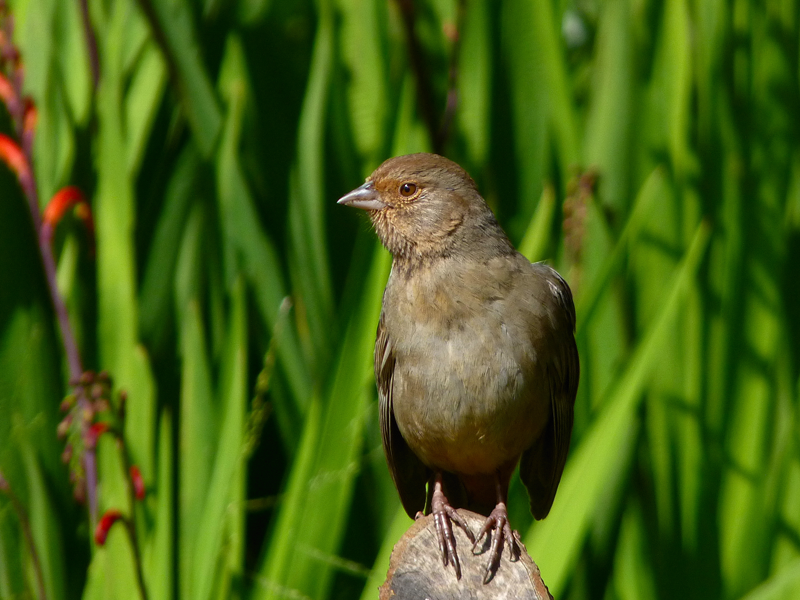}
        \includegraphics[width=0.32\columnwidth]{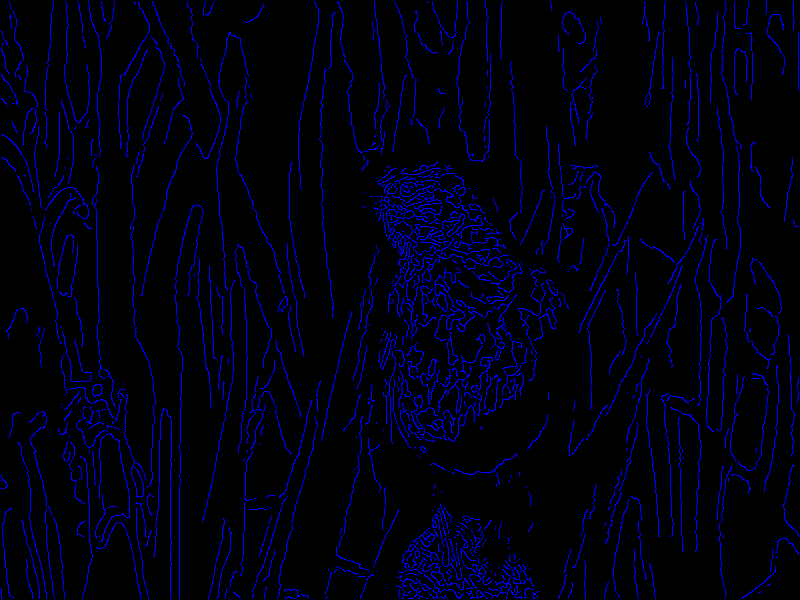}
        \includegraphics[width=0.32\columnwidth]{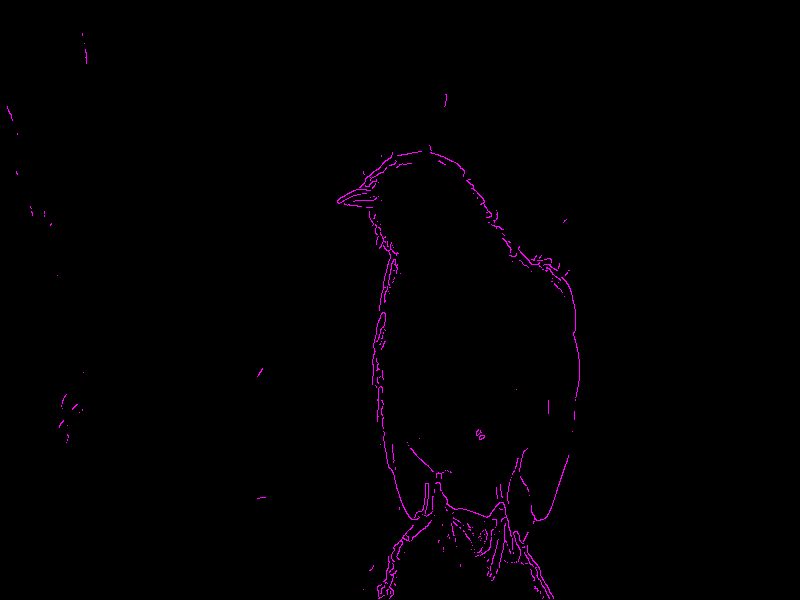}
        \includegraphics[width=0.32\columnwidth]{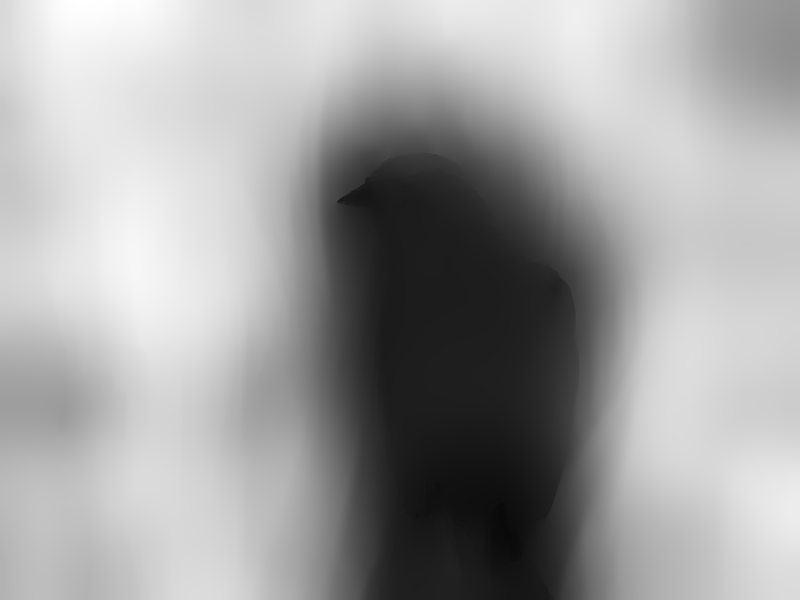}
        \includegraphics[width=0.32\columnwidth]{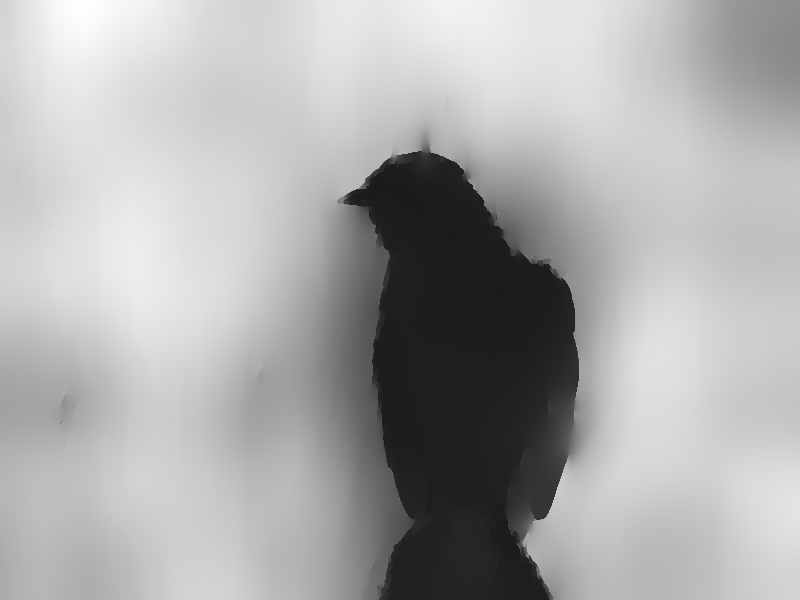}
        \includegraphics[width=0.32\columnwidth]{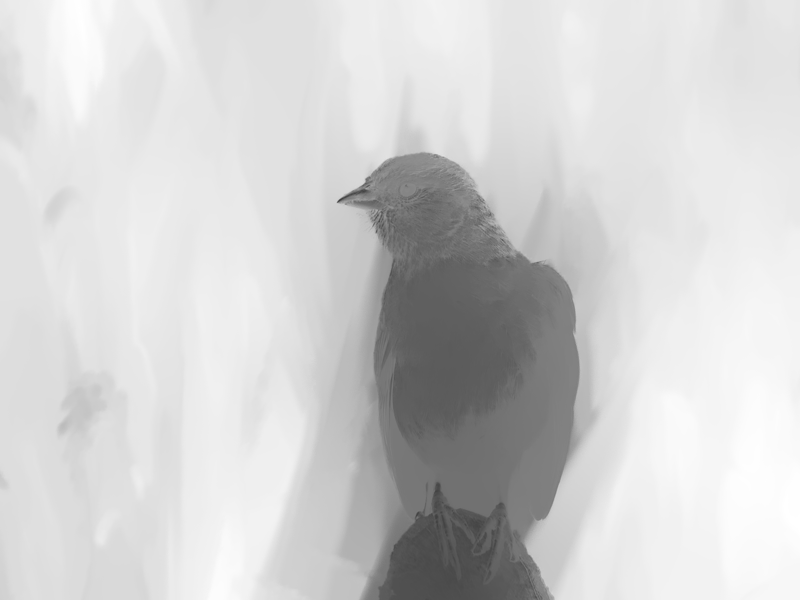}
        \vskip .05cm
        \includegraphics[width=0.32\columnwidth]{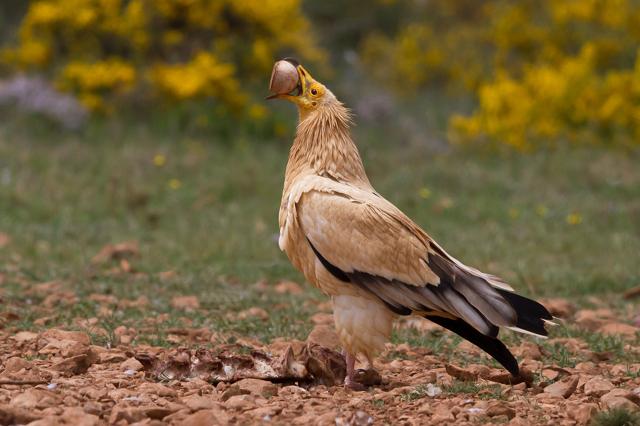}
        \includegraphics[width=0.32\columnwidth]{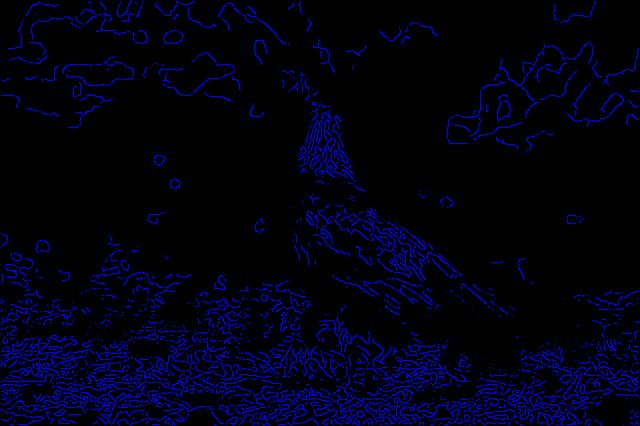}
        \includegraphics[width=0.32\columnwidth]{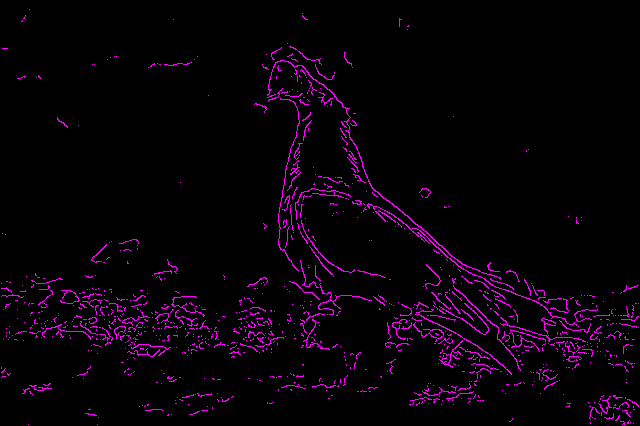}
        \includegraphics[width=0.32\columnwidth]{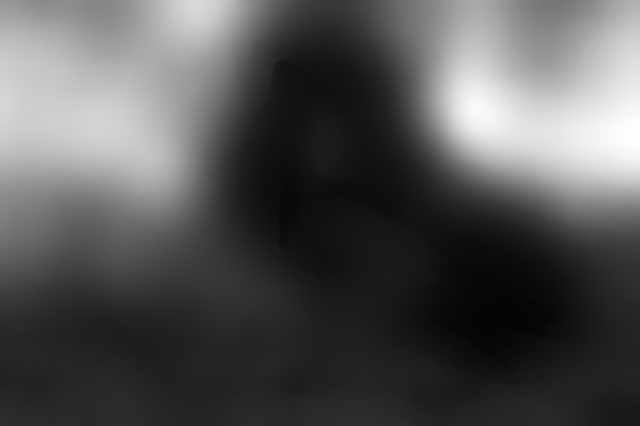}
        \includegraphics[width=0.32\columnwidth]{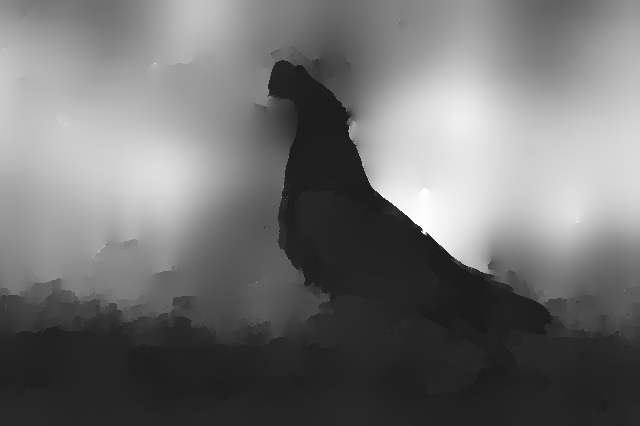}
        \includegraphics[width=0.32\columnwidth]{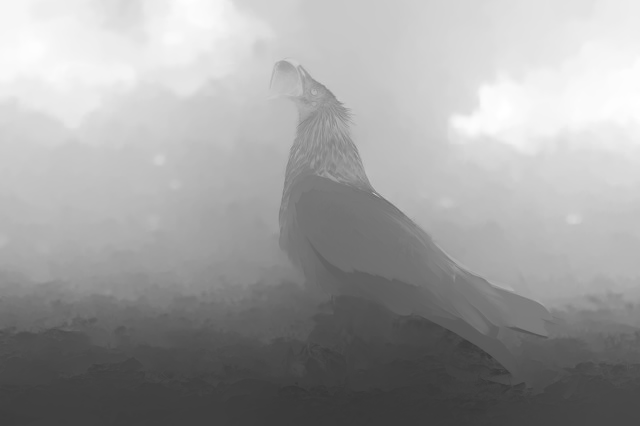}
        \vskip .05cm
        \includegraphics[width=0.32\columnwidth]{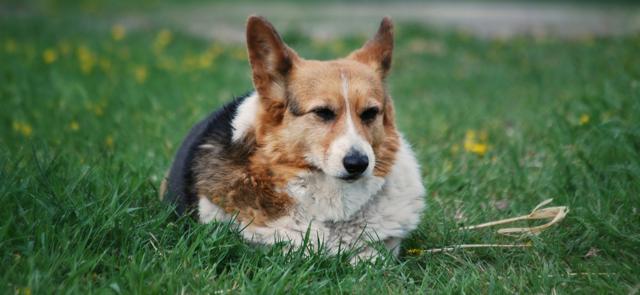}
        \includegraphics[width=0.32\columnwidth]{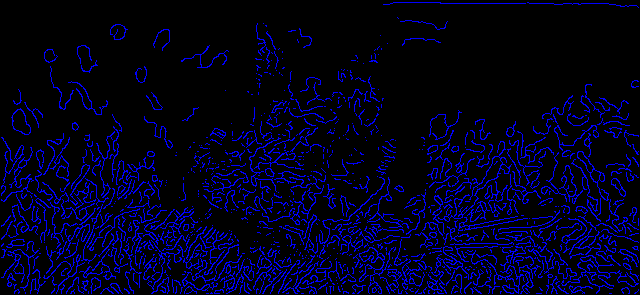}
        \includegraphics[width=0.32\columnwidth]{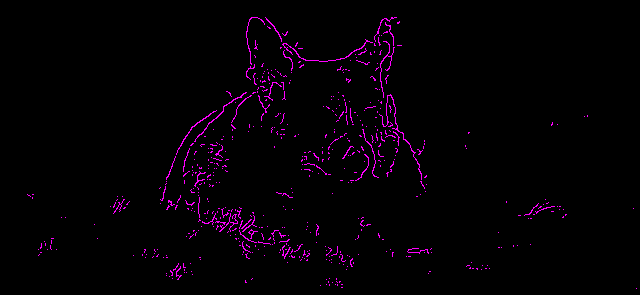}
        \includegraphics[width=0.32\columnwidth]{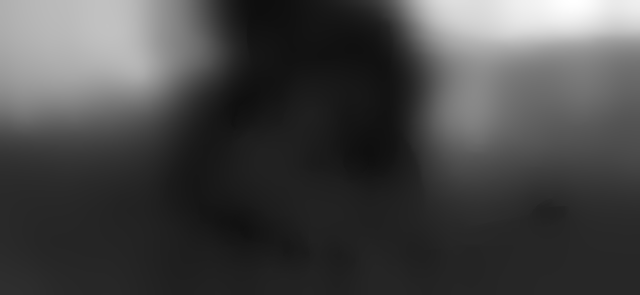}
        \includegraphics[width=0.32\columnwidth]{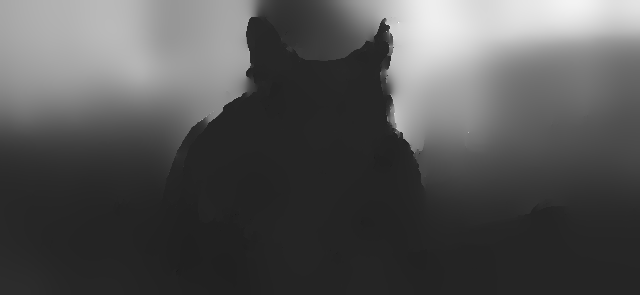}
        \includegraphics[width=0.32\columnwidth]{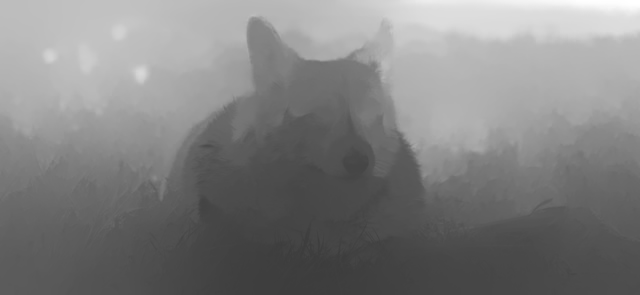}
        \caption{\label{fig:extra1} Columns 1-3: original image, pattern edges, and depth edges. Columns 4-5: full blur maps via DT-Propagation without and with \textit{depth} edge aware propagation. Last column: full blur map via Laplacian-based propagation}
    \end{center}
    \vskip -0.75cm
\end{figure*} 

Fig.~\ref{fig:prcurve} shows the precision-recall curves for our approach and competitive methods, along with the corresponding accuracy (see legend of the figure). Since only a subset of 200 images was used to test the method presented in~\cite{Lee_2019_CVPR} (the remaining 504 were used to train it), all results refer to this smaller test subset for a fair comparison.
It is important to mention that the proposed method, as well as~\cite{DefocusPaper}, \cite{OurTIP2018}, and \cite{park2017unified}, does not use any images from CUHK dataset at any part of the algorithm design or training, while \cite{Lee_2019_CVPR} uses images from CUHK dataset in the training phase for domain adaptation. As an illustration, the last two rows of Fig.~\ref{fig:extra1} show some results produced by our method on images from the CUHK dataset (before thresholding), and more results are provided in the supplementary material.

\subsection{Parameter Settings and Ablation Studies}

This section studies the effect of adding/removing some steps of the proposed approach, as well as the effect of changing some parameters. 
We start by noting that most of the existing edge-based defocus blur estimation methods adopt a post-processing scheme on the sparse blur estimates in order to smooth the results and get rid of outliers. For instance, Zhuo and Sim~\cite{DefocusPaper} used a joint bilateral filter with the original input image as the reference; Karaali and Jung~\cite{OurTIP2018} proposed a Connected Edge Filter (CEF) scheme that regularizes the blur estimates along connected edges; while Park et al.~\cite{park2017unified} proposed a probabilistic joint bilateral filter (PJBF), modifying the post-processing approach used in~\cite{DefocusPaper} by including the confidence values of the blur estimates provided by their network. We tested all these three options in our sparse blur map, and although some of them yielded a small MAE drop on the sparse estimates, none provided any significant error changes in the final full blur maps. More precisely, the error (\ali{raw blur MAE}) of sparse estimates decreased from $0.112$ to $0.110$ when used CEF post-filtering, but increased to $0.116$ when used PJBF. On the other hand, the \ali{raw blur MAE} of the corresponding full blur maps is $0.169$ and $0.171$, respectively, which is equivalent to the results without any post-processing reported in Table~\ref{table:1}.

Another important point to be addressed is the importance of distinguishing \textit{pattern} from \textit{depth} edges, and its relationship with the propagation scheme required
to obtain the full blur map. In Table~\ref{table:1}, we showed that our average  \ali{raw blur MAE} for the natural blur dataset was 0.169. We have also tested our propagation scheme without penalization across \textit{depth} edges, and with the popular Laplacian-based colorization scheme~\cite{LaplacianM} used in several competitive approaches~\cite{DefocusPaper,karaalijungICIP2016,TIP20162,park2017unified}. The average  \ali{raw blur MAE} for these two propagation schemes are 0.188, and 0.288, respectively. We show in Fig.~\ref{fig:extra1} the result from \textit{\textbf{E-NET}} and its impact on  the dense blur map with and without propagation penalization across \textit{depth} edges, along with the Laplacian-based scheme, for three images taken from \cite{DefocusPaper} and \cite{cuhkdata}.  Note that our scheme presents sharper object boundaries in the blur map for the in-focus objects. Also, note that not penalizing depth edges means setting $\psi = 0$ in  Eq.~\eqref{eq:domain_transform_modified}, so that changing $\psi$ from zero to 100 (default value) leads to increasing blockage of blur propagation across depth edges.

\section{Conclusions}

In this work, we have introduced a novel deep edge-based defocus blur estimation method. The main contributions of the proposed method were: the introduction of a network that distinguishes \textit{pattern} and \textit{depth} edges; use of only \textit{pattern} edge points for blur estimation via another neural network architecture to avoid CoC ambiguity; and penalizing the propagation of sparse blur values at \textit{depth} edge points to avoid mixing blur values from objects at different depths.

Quantitative results using the traditional MAE metric of raw \ali{and relative} defocus blur estimates showed that the proposed blur estimation method produced more accurate results for natural blurry images than all the tested SOTA approaches. Furthermore, experiments involving image deblurring based on the estimated blur maps showed that our method outperformed competitive approaches in terms of the PSNR and SSIM of deblurred images. Finally, experiments conducted on the related task of defocus blur detection showed that the proposed method gave promising results with good cross-dataset generalization capabilities.  We believe this is a remarkable achievement for an edge-based approach, particularly considering that we generated synthetic training data with a fixed kernel shape (disk) that is only an approximation of the actual blur kernel.

As future work, we plan to use different kernel shapes (e.g., disk and Gaussian) for generating training data with more variability aiming to further improve the generalization of the network. We also intend to develop another deep network that learns how to adequately propagate sparse blur information to the whole image given the original input color image.

%

%
%

\section*{Acknowledgment}

This work was partly funded by the ADAPT Centre for Digital Content Technology, which is funded under the SFI Research Centres Programme (Grant 13/RC/2016) and is co-funded by the European Regional Development Fund, and also partly supported by Brazilian agencies CAPES (Finance Code 001) and CNPq.


\bibliographystyle{IEEEtran}
\bibliography{tip2020_final}

%








\end{document}